%% file: main.tex
\documentclass[11pt]{article}

\usepackage[preprint]{acl}

\usepackage{times}
\usepackage{latexsym}

\usepackage[T1]{fontenc}

\usepackage[utf8]{inputenc}

\usepackage{microtype}

\usepackage{inconsolata}

\usepackage{graphicx}

\usepackage{algorithm}
\usepackage[noend]{algpseudocode}

\usepackage{booktabs}
\usepackage{multirow}
\usepackage{graphicx}
\usepackage[table]{xcolor}
\usepackage{float}
\usepackage{placeins}

\usepackage{amsmath}

\title{Funnel of Thoughts: Efficient Test-Time Scaling \\ via Early Voting and Rollout Pruning}

\author{
  \textbf{Chanhee Park},
  Sungbin Han,
  Jeongho Yoon,
  Seongtae Hong,
  Heuiseok Lim\thanks{Corresponding author.} \\
  Department of Computer Science and Engineering, Korea University \\
  \texttt{\{pch7678,sungbinhan9039,aa007878,ghdchlwls123,limhseok\}@korea.ac.kr} 
}

\begin{document}
\maketitle
\begin{abstract}
\input{sections/0.Abstract}
\end{abstract}

\input{sections/1.Introduction}

\input{sections/2.Related_work}

\input{sections/3.FoT}

\input{sections/4.Experiment}

\input{sections/5.Conclusion}

\input{sections/6.Limitations}



\bibliography{custom}

\clearpage
\appendix
\input{sections/x.Appx}

\end{document}

%% file: sections/0.Abstract.tex
Large Reasoning Models produce diverse, sometimes inconsistent answers across repeated queries on the same problem, so multi-sample inference is a prerequisite for reliable deployment. Majority voting at $k$ rollouts is the standard solution and the de facto accuracy target for this regime, but it is prohibitively expensive at the scale LRMs require. We introduce \textbf{Funnel of Thoughts} (FoT), an inference-time method that preserves the full 32-trajectory voted accuracy while halving its attention FLOPs, a 28.8\% reduction in full-model inference cost. Across 115K reasoning trajectories from six LRMs, we find that unproductive trajectories often reveal themselves through repeated hesitation markers such as ``Wait,'' ``Actually,'' and ``perhaps.'' These trajectories are less likely to reach the correct answer and consume disproportionate attention FLOPs, degenerating into no-answer loops in the worst case. Built on this training-free lexical signal, FoT identifies the vocabulary that captures these pathological patterns and prunes affected trajectories before completion, reducing online generation attention FLOPs by 56.1\% and wall time by 37.6\% without any additional model inference; the same signal transfers without retuning across held-out architectures and out-of-domain tasks.

%% file: sections/1.Introduction.tex
\section{Introduction}

\label{sec:intro}
Large Reasoning Models (LRMs)~\citep{openai2024openaio1card, Guo_2025, qwq32b} solve problems by generating long chains of thought~\citep{wei2023chainofthoughtpromptingelicitsreasoning}. Compared with Instruct models~\citep{ouyang2022traininglanguagemodelsfollow}, they explore richer answer trajectories and therefore produce diverse, sometimes inconsistent answers across repeated queries~\citep{brown2024largelanguagemonkeysscaling}. This diversity makes multi-sample inference valuable: the correct answer often appears somewhere in the rollout pool, even when no single rollout is reliable. Self-Consistency (SC@$k$; \citealp{wang2023selfconsistencyimproveschainthought}) exploits this by sampling $k$ rollouts and returning the majority vote, routinely recovering 20--40 percentage points over single-rollout accuracy on competition benchmarks~\citep{snell2024scalingllmtesttimecompute}. The cost is that every rollout must run to completion, and LRM rollouts are long, running to thousands of tokens each. Because attention cost grows with the square of the sequence length, every additional token is more expensive than the one before it, so the tail of a long trajectory, rather than its beginning, dominates what SC@$k$ actually pays for. When some trajectories spiral into repetitive, counterproductive self-correction~\citep{wang2025thoughtsplaceunderthinkingo1like, su2025underthinkingoverthinkingempiricalstudy, huang2024largelanguagemodelsselfcorrect}, SC pays for thousands of wasted tokens at exactly the point where tokens cost the most. Prior analyses likewise find that 40--50\% of LRM tokens can be redundant~\citep{chen2025think23overthinkingo1like}.

\begin{figure}
\centering
 \includegraphics[width=\columnwidth]{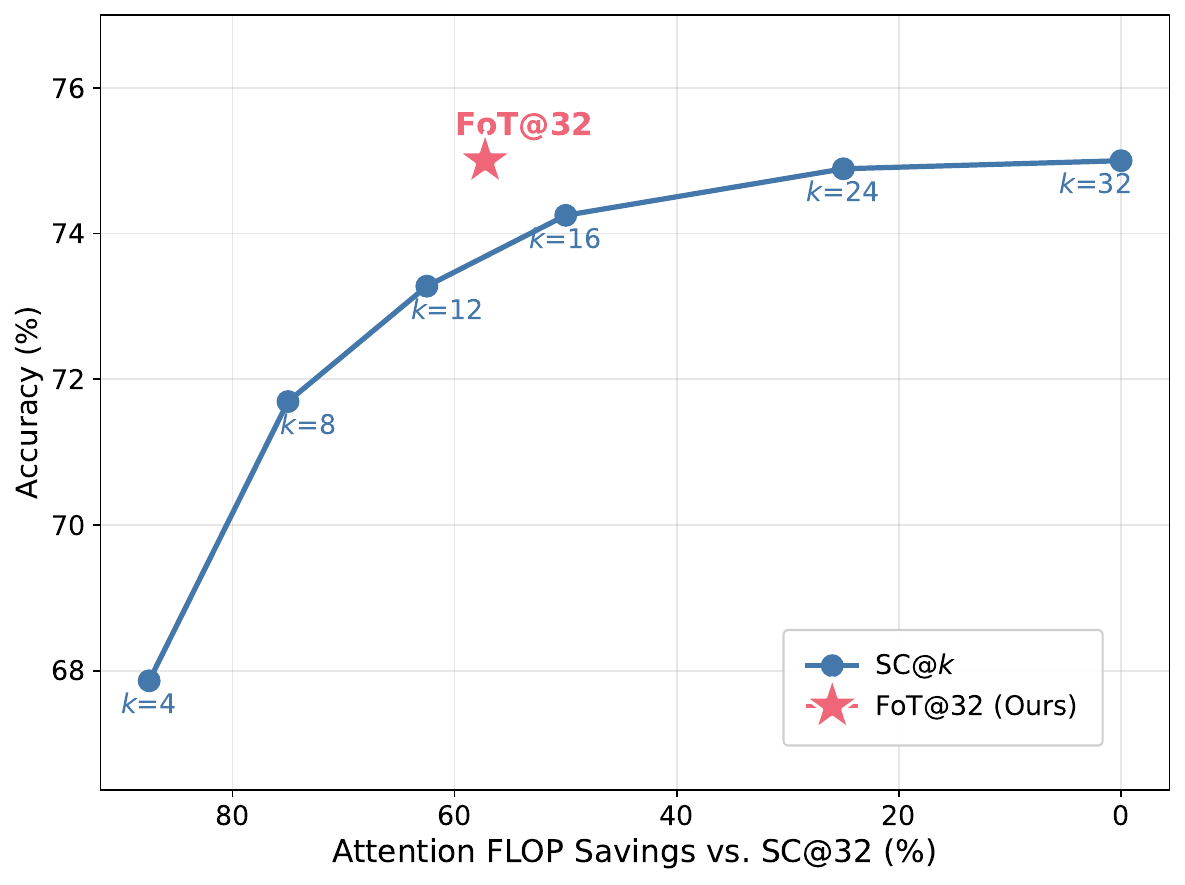}
\caption{Accuracy vs.\ compute on AIME24/25. SC@$k$ trades accuracy for compute by varying pool size $k$. FoT@32 defines a new Pareto point: matching SC@32 accuracy at SC@16-level compute.}
\label{fig:pareto}
\end{figure}

\begin{figure*}[t]
\centering
 \includegraphics[width=0.92\textwidth]{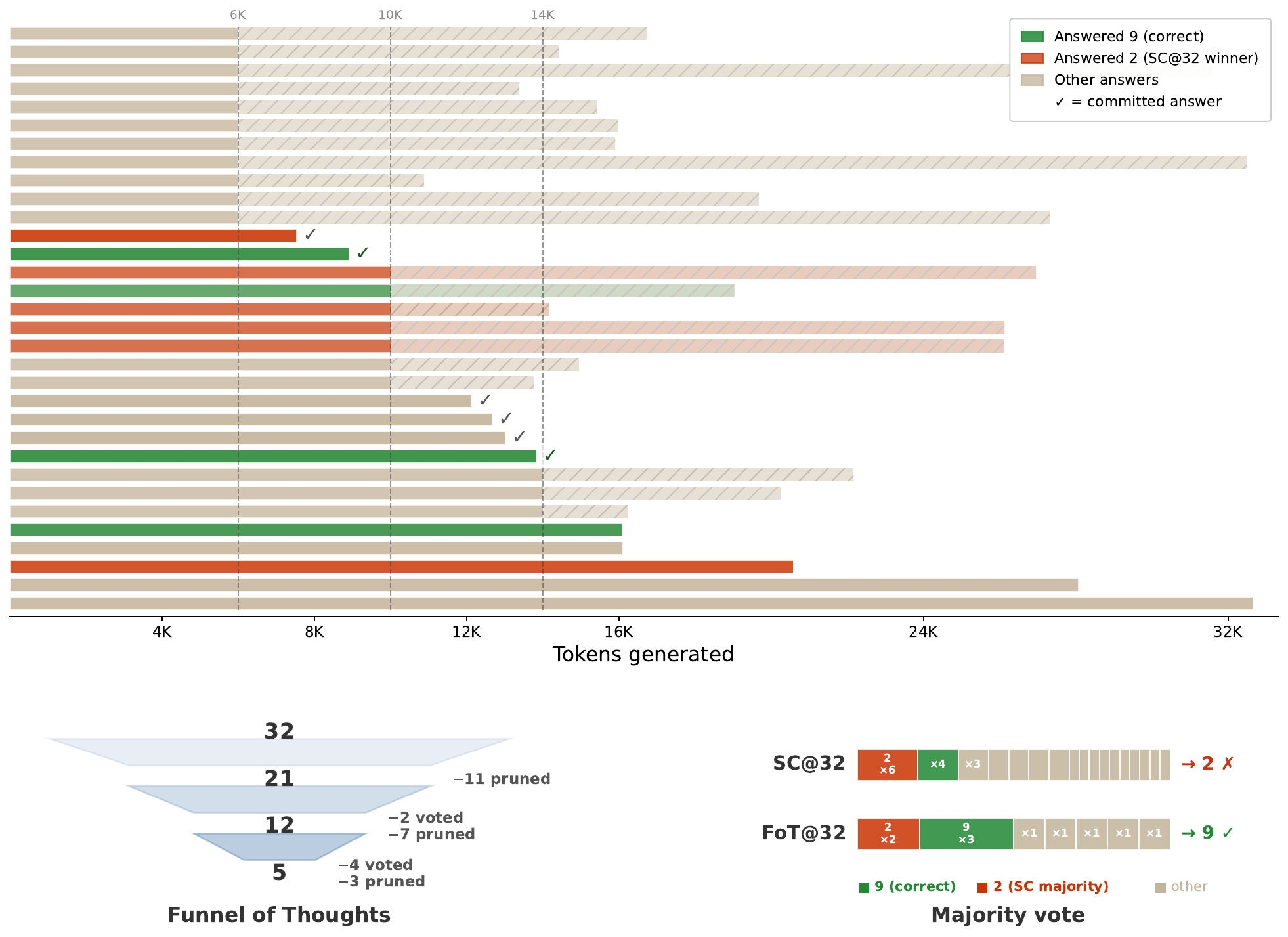}
\caption{Funnel of Thoughts on AMC23 Problem 20 (DeepSeek-R1-Distill-Qwen-7B, $k{=}32$). \textbf{Top:} Each bar is one rollout, ordered by sampling seed, and truncated at its pruning or early-voting checkpoint. \textbf{Bottom-left:} The pool narrows across three checkpoints. \textbf{Bottom-right:} SC@32 elects the wrong plurality (2); FoT@32 lifts the correct answer (9) into the plurality.}
\label{fig:main}
\end{figure*}

To mitigate this inefficiency, we introduce \textbf{Funnel of Thoughts} (FoT), an inference-time algorithm that starts with a full pool of $k$ parallel reasoning trajectories and progressively removes unproductive ones in-flight using only the generated text. Unlike efficient-sampling methods that draw fewer trajectories, FoT runs the full pool in parallel and truncates only its spiraling minority, shedding late-stage waste while keeping the productive trajectories that carry the vote. Our key observation is that hesitation markers, such as ``Wait,'' ``Actually,'' and ``perhaps'', are significantly more frequent in incorrect trajectories. At fixed token-count checkpoints, FoT applies two mechanisms: \textit{early voting} banks trajectories that have already committed to an answer; \textit{rollout pruning} removes those exhibiting the highest hesitation marker density. As Figure~\ref{fig:pareto} demonstrates, FoT preserves SC@32 accuracy on AIME24/25, hard competition-math benchmarks where multi-sample inference is most needed, at a 57\% reduction in attention FLOPs. Figure~\ref{fig:main} traces this on a representative problem, where SC@32's plurality lands on a wrong answer and FoT recovers the correct one.

Our contributions are: (1) a \textbf{data-driven analysis} showing that the density of hesitation markers, a zero-cost lexical signal, identifies the unproductive tail of LRM reasoning across 115K rollouts; (2) \textbf{Funnel of Thoughts}, an inference-time algorithm exploiting this signal to prune rollouts at inference time in parallel batches without training, reward models, or logit access; (3) \textbf{generalizability}: FoT transfers unchanged across architecturally diverse held-out models and out-of-domain tasks, preserving SC@32 accuracy at substantially reduced compute.

%% file: sections/2.Related_work.tex
\section{Related Work}
\label{sec:related}

\subsection{Reasoning Inefficiency in LRMs}

A growing body of work documents systematic inefficiency in LRM reasoning trajectories. \citet{chen2025think23overthinkingo1like} find that about 48\% of generated tokens on MATH500 are spent on redundant operations such as repetitive self-verification and unnecessary case exploration, while \citet{su2025underthinkingoverthinkingempiricalstudy} observe a non-monotonic relationship between trajectory length and accuracy in which both very short and very long rollouts underperform. \citet{huang2024largelanguagemodelsselfcorrect} show that self-correction without external feedback is counterproductive on average. These findings motivate our approach: the self-correction attempts that manifest as hesitation markers are, on aggregate, wasteful.

Building on this diagnosis, a line of work intervenes at the single-trajectory level to suppress unproductive reasoning. TIP \citep{wang2025thoughtsplaceunderthinkingo1like} penalizes excessive thought-switching in logits, \citet{wang2025waitdontneedwait} suppress self-reflection keywords to reduce trajectory length, and \citet{wu2025thoughtcalibrationefficientconfident} use hidden-state probes to dynamically terminate thinking. These approaches focus on shaping the best single rollout. But single-rollout interventions cannot mitigate the fundamental property that LRM inference produces a diverse distribution of final answers, demonstrated by the gap between pass@1 and pass@$k$~\citep{brown2024largelanguagemonkeysscaling}. This diversity grows with trajectory length: every additional decoding step compounds the probability of divergence across rollouts, and LRM trajectories run thousands of tokens.

\subsection{Self-Consistency and Sampling Efficiency}

SC \citep{wang2023selfconsistencyimproveschainthought} samples $k$ reasoning paths and selects the majority vote, yielding large gains but requiring $k$ full-length trajectories. A line of work reduces cost by deciding when to stop sampling: Adaptive Consistency \citep{aggarwal-etal-2023-lets} applies a Beta stopping rule on the running majority vote, Certaindex \citep{fu2025efficientlyscalingllmreasoning} uses a learned stability metric, and others refine the criterion further \citep{wan-etal-2025-reasoning, wang-etal-2025-make}. These methods generate rollouts sequentially: once a trajectory begins, it must run to completion, even when it has entered the wasteful late-stage generation that dominates LRM compute. They act on the sample axis and FoT on the token axis, so the two are complementary; we compare against Adaptive Consistency in our main results and against Difficulty-Adaptive Self-Consistency and Certaindex in Appendix~\ref{app:baselines}.

Parallel alternatives operate within a $k$-rollout pool but pursue different goals. Slim-SC \citep{hong2025slimscthoughtpruningefficient} prunes redundant rollouts during generation via inter-trajectory embedding similarity, requiring a separate embedding model. CISC \citep{Taubenfeld_2025} reweights votes after all rollouts complete using token-level confidence, requiring logprob access and not reducing generation cost. Neither directly targets the pathological trajectories that drive late-stage waste, whereas FoT preemptively identifies and prunes them at inference time using only the generated text.

\subsection{Test-Time Compute Scaling}

More broadly, test-time compute scaling can be as effective as model scaling: optimal allocation lets smaller models match much larger ones \citep{snell2024scalingllmtesttimecompute, wu2025inferencescalinglawsempirical}, and the gap between pass@1 and pass@$k$ on hard problems exceeds 50pp \citep{brown2024largelanguagemonkeysscaling}, motivating the multi-sample methods. Process Reward Models offer strong rollout selection in this regime but require a separately trained verifier \citep{lightman2023letsverifystepstep, wang2024mathshepherdverifyreinforcellms}. The closest prior to our lexical signal is budget forcing \citep{muennighoff2025s1simpletesttimescaling}, which appends \texttt{Wait} tokens to extend reasoning; we use the \emph{density} of the same token, and others like it, as a signal that extended reasoning has become unproductive.

%% file: sections/3.FoT.tex
\section{Funnel of Thoughts}
\label{sec:method}

\subsection{Preliminaries}
\label{sec:prelim}

\paragraph{Pass@$k$.}
Pass@$k$ is the probability that at least one of $k$ rollouts is correct. It represents the upper bound of what the model can achieve given its knowledge, but is not attainable in practice without an oracle: selecting the correct rollout from the pool requires knowing the answer a priori. Reaching Pass@$k$ accuracy at low compute cost is therefore the key challenge for effective multi-sampling; SC@$k$ approximates this ceiling via the majority-vote estimator.

\paragraph{Self-Consistency (SC@$k$).}
Given a problem $q$, SC@$k$ \citep{wang2023selfconsistencyimproveschainthought} independently samples $k$ rollouts $\{c_1, \ldots, c_k\}$ from a language model, extracts a candidate answer $a_i$ from each rollout, and returns the majority answer\footnote{We follow standard SC literature convention in using ``majority'' to denote the most-voted answer regardless of whether it strictly exceeds 50\% of votes; on our hardest problems the top answer is often a plurality rather than a strict majority.}:
\begin{equation}
\hat{a} = \arg\max_{a} \sum_{i=1}^{k} \mathbf{1}[a_i = a].
\end{equation}

\paragraph{Cost metrics.}
Each rollout $c_i$ has total sequence length $s_i$ tokens. Because self-attention is quadratic, we report compute cost primarily as total attention FLOPs across all rollouts: $\text{FLOPs} = \sum_{i=1}^{k} 4 \cdot L \cdot d \cdot s_i^2$, where $L$ is the number of layers and $d$ the model dimension. Attention is the term that grows with trajectory length, and therefore the term that in-flight pruning acts on, which is why we measure it directly. Including feed-forward and projection compute, FoT's saving is 28.8\%, and its end-to-end wall-clock saving on a live server is 37.6\%; we report these in Section~\ref{sec:online_exp} and Appendix~\ref{app:efficient_attention}, and every FLOP figure in this paper states which of the two accountings it uses.

\paragraph{Setup.}
We analyze and evaluate FoT on 6 LRMs: DeepSeek-R1-Distill-Qwen-1.5B and 7B \citep{Guo_2025}, OpenThinker3-7B \citep{guha2025openthoughtsdatarecipesreasoning}, Qwen3-4B-Thinking-2507, Qwen3-30B-A3B-Thinking-2507, and QwQ-32B \citep{yang2025qwen3technicalreport, qwq32b}. The benchmark suite is four competition-math evaluations spanning difficulty levels: AIME24, AIME25, AMC23, and MATH500~\cite{lightman2023letsverifystepstep}. For each problem we generate 32 independent rollouts with unique random seeds, yielding a 115{,}200-rollout pool over 3{,}600 model-problem pairs and nearly 0.8 billion generated tokens. All rollouts are pre-generated; FoT and baselines operate on the same pool. Answers are graded with \texttt{math\_verify}; an earlier draft used exact string match, which depressed only MATH500 absolute values (Appendix~\ref{app:generation}). Regrading moves MATH500 absolute accuracies substantially and can shift an individual cell's FoT--SC gap by a problem or two; the aggregate finding that FoT matches SC@32 is unchanged.

\input{tables/table_observations}

\subsection{Motivating Observations}
\label{sec:observations}

Across the 115{,}200-rollout pool, Table~\ref{tab:observations} summarizes the behavioral profile; three observations motivate our method.

\paragraph{Observation 1: Early commitment predicts correctness.}
Correct rollouts produce a final answer (\texttt{\textbackslash boxed\{\}}) after an average of 5{,}575 tokens, while incorrect rollouts run to 16{,}128. A further 2.1\% of rollouts never commit at all, consuming the maximum budget. Rollouts reaching an answer early have higher chance of being correct, motivating an \textit{early voting} mechanism that preserves committed answers before pruning. This suggests that for problems the model can solve quickly, additional reasoning may not improve and can even degrade the answer~\citep{chen2025think23overthinkingo1like, huang2024largelanguagemodelsselfcorrect}.

\paragraph{Observation 2: Late-stage generation is disproportionately wasteful.}
Rollouts that fail to commit early enter self-correction spirals \citep{huang2024largelanguagemodelsselfcorrect, wang2025thoughtsplaceunderthinkingo1like}. Because attention compute is quadratic in length, these long trajectories are also the expensive ones: an incorrect rollout costs 6.0 times the attention FLOPs of a correct one, and a rollout that never commits costs 17.4 times, despite the two together making up under a tenth of the pool. This asymmetry motivates targeting the wasteful tail via \textit{rollout pruning} rather than uniformly reducing all rollouts.

\paragraph{Observation 3: Hesitation markers signal unproductive reasoning.}
Hesitation markers---``\texttt{Wait,}'', ``\texttt{actually}'', ``\texttt{perhaps}'', and 18 others---are denser in incorrect rollouts across all 6 LRMs. We measure \textit{hesitation marker density} as marker count per 1{,}000 characters, computed cumulatively over the generated text. Sorting rollouts into density deciles yields a monotonic accuracy decline ($r{=}-0.82$ across deciles, $r_{\text{pb}}{=}-0.30$ per rollout; 31pp gap between D1 and D10). Density is the online-available leading indicator of runaway length: it forecasts a rollout's final length from the visible prefix ($r{=}{+}0.13$ to ${+}0.41$ across models) before that length has been paid for, marking the unproductive tail FoT should prune. We report the details of this process in Appendix~\ref{app:motivating}.

\subsection{Algorithm}
\label{sec:algorithm}

\input{algorithms/fot_algo}

As illustrated in Algorithm~\ref{alg:fot}, Funnel of Thoughts retains the core SC@$k$ framework, sampling $k$ rollouts and taking the majority vote, but eliminates unproductive rollouts to save compute. The method maintains an \textit{active set} $\mathcal{A}$ of rollouts still being generated and a \textit{vote bank} $\mathcal{B}$ of answers from rollouts that have already committed. At each of $m$ fixed token-count checkpoints $\{t_1, t_2, \ldots, t_m\}$, two mechanisms are applied. After the final checkpoint, the answer is selected by \textsc{Plurality} (argmax over answer counts) across $\mathcal{B}$ and the surviving rollouts in $\mathcal{A}$. The only hyperparameters are the checkpoint schedule and keep ratio $\rho$, selected via a grid sweep on the pool and held constant across all problems, models, and checkpoints. Detailed results can be found in Appendix~\ref{sec:ablation_sweep}.

\paragraph{Early Voting.}
At checkpoint $t_j$, we inspect the text generated so far for each active rollout $c_i$. Any rollout that has already produced a final answer (\texttt{\textbackslash boxed\{\}}) is moved to the \textit{vote bank} $\mathcal{B}$: its answer is preserved for the final vote, and the rollout is removed from the active set. This is what makes the method a funnel rather than a truncation. Without banking, preserving $k$ votes would require running $k$ trajectories to completion, so narrowing the pool would mean discarding votes; banking separates the two, letting the pool shrink while the ballot stays full. Across our pool, only 3.1 of 32 trajectories are still generating at the final checkpoint, yet the vote is cast by 19.7 of them, where the same funnel without banking would leave 8.4. Banking also confines the pruning signal to the rollouts it applies to, since a committed rollout's outcome is already settled and cannot be predicted by a measure of future spiraling. Early voting requires only substring matching, with no model inference.

\input{tables/table1_main}
\paragraph{Rollout Pruning.}
Among the remaining active rollouts, we compute hesitation marker density $d_i$ over the entire text generated from rollout start to checkpoint $t_j$. Rollouts are ranked by density in ascending order; we retain the $\lfloor |\mathcal{A}| \cdot \rho \rfloor$ rollouts with the lowest hesitation rates and discard the rest, where $\rho \in (0, 1]$ is a fixed keep ratio. We always retain at least 2 active rollouts.

\paragraph{Online Deployment.}
FoT integrates directly into parallel inference pipelines. At each checkpoint, the inference server pauses active rollouts, applies early voting and pruning, and resumes survivors. Terminated rollouts release their KV cache immediately, reducing memory pressure and freeing batch slots for the remaining rollouts. Because the pruning decision requires only the generated text, with no separate embedding model, reward model, or logprob access, FoT adds negligible overhead to the serving loop. We validate this integration in Section~\ref{sec:online_exp}.

%% file: tables/table_observations.tex
\begin{table}[t]
\centering
\resizebox{\columnwidth}{!}{%
\begin{tabular}{lccc}
\toprule
& \textbf{Correct} & \textbf{Incorrect} & \textbf{No Answer} \\
\midrule
Proportion of pool         & 90.5\% & 7.4\% & 2.1\% \\
Avg.\ tokens generated     & 5,575  & 16,128  & 31,319 \\
Hesitation density (/1K chars) & 3.46   & 6.14   & 6.93 \\
Relative attention FLOPs   & 1.0$\times$ & 6.0$\times$ & 17.4$\times$ \\
\bottomrule
\end{tabular}%
}
\caption{Behavioral profile of the 115{,}200-rollout pool across 6 models and 4 benchmarks. Correct rollouts commit early, carry fewer hesitation markers, and cost a fraction of the compute; the 9.5\% of rollouts that end wrong or never commit consume a disproportionate share of it.}
\label{tab:observations}
\end{table}

%% file: algorithms/fot_algo.tex

\begin{algorithm}[t]
\caption{Funnel of Thoughts (FoT)}
\label{alg:fot}
\begin{footnotesize}
\begin{algorithmic}[1]
\Require Rollouts $\{c_1, \ldots, c_k\}$, checkpoints $\{t_1, \ldots, t_m\}$, keep ratio $\rho$
  \State $\mathcal{A} \gets \{1, \ldots, k\}$ \Comment{active set}
  \State $\mathcal{B} \gets \emptyset$ \Comment{vote bank}
  \For{$j \gets 1$ \textbf{to} $m$} \Comment{loop over checkpoints}
    \If{$|\mathcal{A}| \leq 2$}
      \State \textbf{break}
    \EndIf
    \For{each $i \in \mathcal{A}$} \Comment{\textit{Early Voting}}
      \If{$\textsc{HasAnswer}(c_i[1{:}t_j])$}
        \State $\mathcal{B} \gets \mathcal{B} \cup \{\textsc{Extract}(c_i[1{:}t_j])\}$
        \State $\mathcal{A} \gets \mathcal{A} \setminus \{i\}$
      \EndIf
    \EndFor
    \If{$|\mathcal{A}| \leq 2$}
      \State \textbf{continue}
    \EndIf
    \State $d_i \gets \textsc{HesitationDensity}(c_i[1{:}t_j])$ for all $i \in \mathcal{A}$ \Comment{\textit{Rollout Pruning}}
    \State $n \gets \max(2,\; \lfloor |\mathcal{A}| \cdot \rho \rfloor)$
    \State $\mathcal{A} \gets$ indices of $n$ lowest-density rollouts
  \EndFor
  \State Complete generation for all $i \in \mathcal{A}$
  \State \textbf{return} $\textsc{Plurality}(\mathcal{B} \cup \{\textsc{Extract}(c_i) : i \in \mathcal{A}\})$
\end{algorithmic}
\end{footnotesize}
\end{algorithm}

%% file: tables/table1_main.tex
\providecommand{\aUp}[1]{\cellcolor{green!16}#1}  
\providecommand{\aDn}[1]{\cellcolor{red!26}#1}    
\providecommand{\aEq}[1]{#1}                       
\begin{table*}[t]
\centering
\footnotesize
\renewcommand{\arraystretch}{0.95}
\setlength{\tabcolsep}{5pt}
\begin{tabular}{ll cc cc cc cc cc}
\toprule
& & \multicolumn{2}{c}{\textbf{AIME24}} & \multicolumn{2}{c}{\textbf{AIME25}} & \multicolumn{2}{c}{\textbf{AMC23}} & \multicolumn{2}{c}{\textbf{MATH500}} & \multicolumn{2}{c}{\textbf{Overall (/600)}} \\
\cmidrule(lr){3-4} \cmidrule(lr){5-6} \cmidrule(lr){7-8} \cmidrule(lr){9-10} \cmidrule(lr){11-12}
Model & Method & Acc & FLOP & Acc & FLOP & Acc & FLOP & Acc & FLOP & Acc & FLOP \\
\midrule
\multirow{4}{*}{DS-R1-1.5B}
  & SC@32  & 56.7 & --- & 33.3 & --- & 90.0 & --- & 92.4 & --- & 87.5 & --- \\
  & AC  & \aEq{56.7} & \cellcolor{gray!4}10.4\% & \aEq{33.7} & \cellcolor{gray!4}12.7\% & \aEq{90.0} & \cellcolor{gray!8}39.4\% & \aEq{92.2} & \cellcolor{gray!8}43.3\% & \aEq{87.3} & \cellcolor{gray!4}26.4\% \\
  & SlimSC  & \aUp{63.3} & \cellcolor{gray!16}56.3\% & \aUp{40.0} & \cellcolor{gray!24}68.7\% & \aUp{92.5} & \cellcolor{gray!24}60.3\% & \aEq{91.8} & \cellcolor{gray!16}54.5\% & \aEq{87.8} & \cellcolor{gray!24}60.0\% \\
  & \textbf{FoT@32}  & \aUp{60.0} & \cellcolor{gray!16}57.6\% & \aEq{33.3} & \cellcolor{gray!16}54.3\% & \aEq{90.0} & \cellcolor{gray!16}55.9\% & \aEq{92.6} & \cellcolor{gray!16}50.3\% & \aEq{87.8} & \cellcolor{gray!16}54.5\% \\
\midrule
\multirow{4}{*}{DS-R1-7B}
  & SC@32  & 80.0 & --- & 53.3 & --- & 95.0 & --- & 95.4 & --- & 92.5 & --- \\
  & AC  & \aEq{80.0} & \cellcolor{gray!8}31.3\% & \aEq{53.3} & \cellcolor{gray!4}28.0\% & \aEq{95.0} & \cellcolor{gray!24}60.8\% & \aEq{95.4} & \cellcolor{gray!16}56.5\% & \aEq{92.5} & \cellcolor{gray!8}44.1\% \\
  & SlimSC  & \aUp{83.3} & \cellcolor{gray!16}53.8\% & \aEq{53.3} & \cellcolor{gray!24}64.9\% & \aUp{97.5} & \cellcolor{gray!8}41.0\% & \aEq{95.2} & \cellcolor{gray!8}43.5\% & \aEq{92.7} & \cellcolor{gray!16}50.8\% \\
  & \textbf{FoT@32}  & \aEq{80.0} & \cellcolor{gray!16}54.8\% & \aUp{56.7} & \cellcolor{gray!16}56.3\% & \aUp{100.0} & \cellcolor{gray!16}46.6\% & \aEq{95.4} & \cellcolor{gray!8}38.7\% & \aEq{93.0} & \cellcolor{gray!16}49.1\% \\
\midrule
\multirow{4}{*}{OT3-7B}
  & SC@32  & 76.7 & --- & 73.3 & --- & 97.5 & --- & 97.2 & --- & 95.0 & --- \\
  & AC  & \aEq{76.7} & \cellcolor{gray!16}55.5\% & \aEq{73.3} & \cellcolor{gray!16}49.9\% & \aEq{97.5} & \cellcolor{gray!24}67.3\% & \aEq{97.2} & \cellcolor{gray!24}71.7\% & \aEq{95.0} & \cellcolor{gray!24}61.1\% \\
  & SlimSC  & \aDn{73.3} & \cellcolor{gray!32}77.0\% & \aDn{50.0} & \cellcolor{gray!32}81.6\% & \aEq{97.5} & \cellcolor{gray!24}69.5\% & \aDn{95.8} & \cellcolor{gray!16}58.0\% & \aDn{92.5} & \cellcolor{gray!24}71.5\% \\
  & \textbf{FoT@32}  & \aUp{80.0} & \cellcolor{gray!24}60.3\% & \aUp{76.7} & \cellcolor{gray!24}60.8\% & \aUp{100.0} & \cellcolor{gray!16}53.8\% & \aEq{97.2} & \cellcolor{gray!8}43.6\% & \aEq{95.5} & \cellcolor{gray!16}54.6\% \\
\midrule
\multirow{4}{*}{Qwen3-4B}
  & SC@32  & 86.7 & --- & 86.7 & --- & 100.0 & --- & 98.4 & --- & 97.3 & --- \\
  & AC  & \aEq{86.7} & \cellcolor{gray!16}57.9\% & \aEq{86.7} & \cellcolor{gray!16}48.4\% & \aEq{100.0} & \cellcolor{gray!32}86.9\% & \aEq{98.4} & \cellcolor{gray!24}73.3\% & \aEq{97.3} & \cellcolor{gray!24}66.6\% \\
  & SlimSC  & \aDn{76.7} & \cellcolor{gray!24}65.7\% & \aDn{80.0} & \cellcolor{gray!24}61.9\% & \aDn{95.0} & \cellcolor{gray!8}43.7\% & \aEq{98.0} & \cellcolor{gray!16}51.7\% & \aDn{95.8} & \cellcolor{gray!16}55.8\% \\
  & \textbf{FoT@32}  & \aEq{86.7} & \cellcolor{gray!16}58.0\% & \aDn{80.0} & \cellcolor{gray!16}59.4\% & \aEq{100.0} & \cellcolor{gray!16}45.3\% & \aEq{98.4} & \cellcolor{gray!8}43.8\% & \aEq{97.0} & \cellcolor{gray!16}51.6\% \\
\midrule
\multirow{4}{*}{Qwen3-30B}
  & SC@32  & 93.3 & --- & 86.7 & --- & 100.0 & --- & 98.0 & --- & 97.3 & --- \\
  & AC  & \aEq{93.3} & \cellcolor{gray!24}72.2\% & \aUp{88.3} & \cellcolor{gray!16}56.8\% & \aEq{100.0} & \cellcolor{gray!32}87.3\% & \aEq{97.9} & \cellcolor{gray!24}71.7\% & \aEq{97.3} & \cellcolor{gray!24}72.0\% \\
  & SlimSC  & \aEq{93.3} & \cellcolor{gray!8}42.1\% & \aDn{83.3} & \cellcolor{gray!16}49.9\% & \aEq{100.0} & \cellcolor{gray!4}29.4\% & \aEq{98.0} & \cellcolor{gray!8}35.8\% & \aEq{97.2} & \cellcolor{gray!8}39.3\% \\
  & \textbf{FoT@32}  & \aEq{93.3} & \cellcolor{gray!16}54.5\% & \aDn{83.3} & \cellcolor{gray!16}59.0\% & \aEq{100.0} & \cellcolor{gray!8}38.4\% & \aEq{98.2} & \cellcolor{gray!8}37.7\% & \aEq{97.3} & \cellcolor{gray!16}47.4\% \\
\midrule
\multirow{4}{*}{QwQ-32B}
  & SC@32  & 90.0 & --- & 83.3 & --- & 100.0 & --- & 97.4 & --- & 96.5 & --- \\
  & AC  & \aEq{90.0} & \cellcolor{gray!24}61.2\% & \aEq{83.3} & \cellcolor{gray!16}53.8\% & \aEq{100.0} & \cellcolor{gray!32}84.7\% & \aEq{97.4} & \cellcolor{gray!24}74.5\% & \aEq{96.5} & \cellcolor{gray!24}68.5\% \\
  & SlimSC  & \aDn{83.3} & \cellcolor{gray!24}69.6\% & \aDn{63.3} & \cellcolor{gray!32}75.5\% & \aEq{100.0} & \cellcolor{gray!16}50.4\% & \aEq{97.2} & \cellcolor{gray!16}45.7\% & \aDn{95.0} & \cellcolor{gray!24}60.3\% \\
  & \textbf{FoT@32}  & \aDn{86.7} & \cellcolor{gray!16}54.0\% & \aEq{83.3} & \cellcolor{gray!16}58.1\% & \aEq{100.0} & \cellcolor{gray!8}42.8\% & \aEq{97.2} & \cellcolor{gray!8}33.5\% & \aEq{96.2} & \cellcolor{gray!16}47.1\% \\
\specialrule{0.08em}{0.4em}{0.2em}
\multirow{4}{*}{\textbf{Avg.}}
  & SC@32  & 80.6 & --- & 69.4 & --- & 97.1 & --- & 96.5 & --- & 94.4 & --- \\
  & AC  & \aEq{80.6} & \cellcolor{gray!16}48.1\% & \aEq{69.8} & \cellcolor{gray!8}41.6\% & \aEq{97.1} & \cellcolor{gray!24}71.1\% & \aEq{96.4} & \cellcolor{gray!24}65.2\% & \aEq{94.3} & \cellcolor{gray!16}56.5\% \\
  & SlimSC  & \aDn{78.9} & \cellcolor{gray!24}60.8\% & \aDn{61.7} & \cellcolor{gray!24}67.1\% & \aEq{97.1} & \cellcolor{gray!16}49.1\% & \aEq{96.0} & \cellcolor{gray!16}48.2\% & \aEq{93.5} & \cellcolor{gray!16}56.3\% \\
  & \textbf{FoT@32}  & \aEq{81.1} & \cellcolor{gray!16}56.5\% & \aEq{68.9} & \cellcolor{gray!16}58.0\% & \aUp{98.3} & \cellcolor{gray!16}47.1\% & \aEq{96.5} & \cellcolor{gray!8}41.3\% & \aEq{94.5} & \cellcolor{gray!16}50.7\% \\
\bottomrule
\end{tabular}
\caption{Main results across 6 models and 4 competition-math benchmarks ($k{=}32$). Acc is accuracy (\%) and FLOP the attention-FLOP saving over SC@32, with Overall FLOP taken as the per-benchmark mean so that MATH500's size does not dominate. Accuracy cells are shaded relative to SC@32 (\colorbox{green!16}{higher}, \colorbox{red!26}{lower}, unshaded within $\pm1$pp) and FLOP cells \colorbox{gray!16}{grey}-shaded by size of saving.}
\label{tab:main}
\end{table*}

%% file: sections/4.Experiment.tex
\section{Experiments}
\label{sec:experiments}

Using the same 6-model, 4-benchmark setup from Section~\ref{sec:method}, we compare FoT against SC@32 and two efficient-SC baselines: Adaptive Consistency (AC; \citealp{aggarwal-etal-2023-lets}) and Slim-SC \citep{hong2025slimscthoughtpruningefficient}. We then test generalization to held-out models, cross-domain benchmarks, and online deployment, and ablate the contribution of each FoT mechanism. Sampling parameters, baseline configurations, and the full hyperparameter sweep are in Appendix~\ref{app:implementation}.

\subsection{Main Results}
\label{sec:main}

\paragraph{FoT preserves SC@32 accuracy at half the attention FLOPs.} Across the 4 math benchmarks, FoT@32 preserves the accuracy of full self-consistency while reducing attention FLOPs by roughly half. FoT@32 and SC@32 return the same correctness outcome on 3,580 of 3,600 paired model-problem instances, differing on only 20 cases: 12 FoT wins and 8 FoT losses.\footnote{The difference is not significant under McNemar's exact test ($p{=}0.50$), with a paired-bootstrap 95\% CI on the accuracy difference of $[-0.14,+0.36]$pp.} As Table~\ref{tab:main} demonstrates, FoT holds its FLOP saving in nearly every cell with accuracy preserved, making it a stable operating point on the accuracy--compute tradeoff. It starts from the same 32-rollout pool as SC@32, banks rollouts that have already committed to an answer, and terminates only the active trajectories whose generated text indicates unproductive hesitation. The savings therefore come from removing late-stage waste, not from giving up the answer diversity that makes multi-sample inference useful.

\input{tables/table_passk_regime}

\paragraph{AC struggles when rollouts diverge.} Adaptive Consistency reduces cost by deciding when enough completed rollouts have been sampled. This is effective when the model reaches answer consensus quickly, but the same condition also means SC@32 was over-provisioned for that benchmark. The pass@1--pass@32 gap reveals the difficulty split as a factor: when the gap is large, the correct answer is often present somewhere in the rollout pool, but completed rollouts do not agree quickly enough for sample-axis early stopping. Table~\ref{tab:passk_regime} shows that AC is strongest on easier cells with fast consensus, whereas FoT continues to save compute in the hard split because its decision is made within each rollout. Compared with AC, FoT is more robust on hard samples where consensus is slow but individual trajectories still reveal pruning signals. The two axes trade places with difficulty: sample-axis savings shrink as consensus slows, while FoT's grow. Because the axes are independent, the methods compose rather than compete, which we report in Appendix~\ref{app:baselines}.

\paragraph{Slim-SC over-prunes when faced with hard samples.} Slim-SC exposes a different failure mode: pruning by inter-trajectory similarity can discard useful diversity. This weakness is most visible on AIME25, where Slim-SC saves 67.1\% FLOPs but drops the average accuracy from 69.4 to 61.7 as shown in Table~\ref{tab:main}. On hard problems, correct solutions often share surface structure even when they provide independently useful evidence for the final vote, so treating similar thought segments as redundant can collapse the pool before enough support accumulates. FoT targets a different property: its hesitation-density signal asks whether a trajectory is spiraling, not whether it resembles another trajectory. Compared with Slim-SC, FoT better preserves the reasoning diversity that gives multi-sample inference its advantage.
\begin{figure}[H]
\centering
\includegraphics[width=\columnwidth]{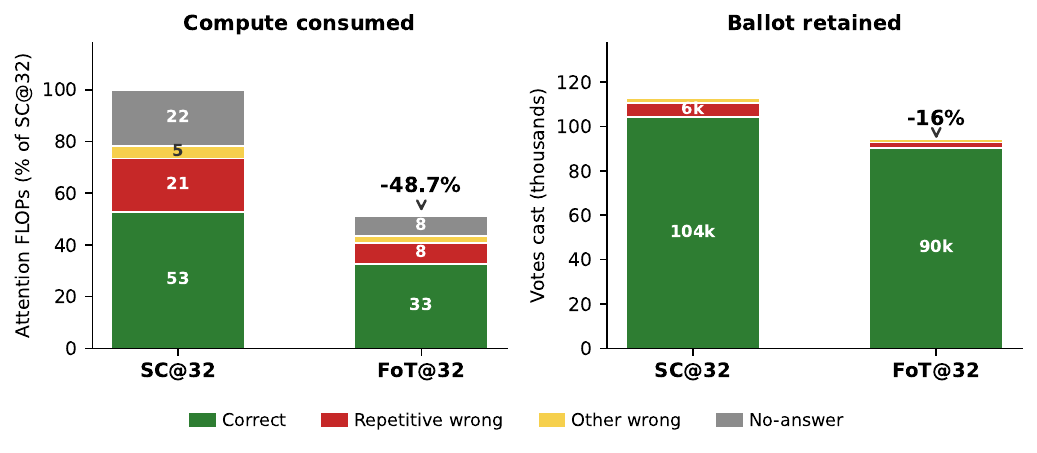}
\caption{\textbf{FoT removes compute, not votes.} Left: attention FLOPs actually consumed, as a percentage of SC@32's total, so a pruned rollout is charged for everything it generated before termination. Right: votes cast in the final plurality.}
\label{fig:typology}
\end{figure}

\paragraph{FoT prunes the failure mode it was designed to catch.} Figure~\ref{fig:typology} separates what pruning costs from what it saves. Repetitive-wrong and no-answer trajectories together consume 42\% of SC@32's attention FLOPs while casting 6\% of its votes, and FoT removes about three fifths of that compute. Correct trajectories keep 87\% of their votes, and the pool still casts 84\% of the original ballot at half the compute, because banking preserves the vote of a rollout that has stopped generating. A checkpoint-level diagnostic gives the same picture for the non-committing tail: Table~\ref{tab:no_answer_reduction} shows that FoT removes 71\% of it by the final checkpoint, with the largest reductions in cells where the issue is most acute. Hesitation-density pruning therefore does not reshape the answer distribution uniformly; it preferentially removes the long repetitive loops~\citep{wang2025thoughtsplaceunderthinkingo1like, su2025underthinkingoverthinkingempiricalstudy} that consume disproportionate attention cost while contributing few votes.


\subsection{Cross-Model Generalization}
\label{sec:cross_model}

FoT transfers across model architectures because its pruning rule is relative rather than model-calibrated. We apply the same token set, checkpoints, and keep ratio to four held-out models: Qwen3-14B, Qwen3-32B, Skywork-OR1-7B \citep{skywork-or1-2025}, and Phi-4-reasoning \citep{abdin2025phi4reasoningtechnicalreport}. These models differ in their absolute hesitation-marker density (Table~\ref{tab:model_density}), but FoT only compares rollouts within the same active pool at the same checkpoint. A model can therefore be more or less verbose overall; what matters is whether a trajectory is unusually hesitation-heavy relative to its peers. Table~\ref{tab:heldout} shows that this criterion preserves SC@32 accuracy with comparable FLOP savings across the held-out models. Phi-4-reasoning is the most architecturally distinct case and also shows the largest transfer gain, further supporting the interpretation. We interpret this gain as arising from Phi-4-reasoning's heavy no-answer tail, reported in Table~\ref{tab:phi4_pathology}, which FoT can capture through the same relative pruning rule.

\input{tables/table3_heldout}


\subsection{Cross-Domain Generalization}
\label{sec:cross_domain}

We next ask whether the same signal survives when the answer format changes. GPQA-Diamond \citep{rein2023gpqagraduatelevelgoogleproofqa} changes the knowledge domain and final-answer format while retaining natural-language deliberation; LCB-Lite \citep{jain2024livecodebenchholisticcontaminationfree} is a stronger stress test because the final artifact is executable code and two correct solutions may be syntactically unrelated. If FoT were exploiting math-specific answer syntax or majority-vote structure, we should not expect it to help under execution-based scoring. Following standard practice for code generation, we evaluate LCB by pass@$k$ over the rollout pool, reported alongside the mean per-rollout pass rate. Early voting also needs a notion of commitment in each domain: on GPQA a rollout commits when it emits a final option letter, and on LCB when it closes a code block after its reasoning. The pruning signal itself is unchanged, since it reads only the deliberation text and never the answer. Table~\ref{tab:cross_domain} shows that the unchanged configuration transfers to both settings, and on LCB FoT improves mean per-rollout pass rate while reducing FLOPs; pass@$k$ declines modestly with $-2.8$pp on average, the expected cost of any method that shrinks the oracle pool. The implication is that hesitation density acts before answer extraction: it identifies low-value continuations in the reasoning process, whether the final output is a boxed number, a multiple-choice option, or code.

\input{tables/table_cross_domain}

\subsection{Online Deployment}
\label{sec:online_exp}

We deploy FoT in an online inference pipeline using SGLang \citep{zheng2024sglangefficientexecutionstructured} on a single A100 GPU, generating all rollouts end-to-end for 30 AIME24 pro
blems with DeepSeek-R1-Distill-Qwen-7B. Table~\ref{tab:online} shows that FoT@32 solves 24 of 30 problems versus 23 for SC@32, while reducing wall time by 37.6\% and attention FLOPs by 56.1\%. The same logs show that the cumulative generated-token footprint, and hence KV-cache occupancy, drops from 12.88M to 8.51M tokens. The wall-time savings follow directly from pruning: terminated rollouts release KV cache memory and batch slots, reducing both compute and memory pressure for the remaining rollouts. Because the hesitation marker count is a string operation with negligible overhead, FoT requires no additional model inference beyond the rollouts themselves.

\input{tables/table5_online}

\subsection{Ablations}
\label{sec:ablations}

We ablate two properties of FoT in the main text: the pool size at which it becomes effective, and the contribution of each mechanism. Kernel-size robustness and additional sensitivity analyses are deferred to Appendix~\ref{app:ablations}.

\paragraph{Pool-size scaling.}
FoT is a large-pool method. Sweeping $k$ from 8 to 32 shows that pruning is mildly harmful when the pool is small, becomes neutral around $k{=}16$, and turns positive beyond it: a small pool cannot spare the vote diversity that pruning removes, whereas a large one can (Appendix~\ref{app:sensitivity}). This sets the regime in which FoT should be used, as a replacement for large-pool self-consistency rather than a small-$k$ method.

\input{tables/table7_evrp}

\paragraph{Mechanism ablation.}
Table~\ref{tab:ablation_evrp} reports a two-by-two ablation on the same 4-benchmark pool as the main result: random vs.\ hesitation-density pruning, each with and without early voting, plus SC@32 as the no-pruning baseline. The result is direct: the lexical pruning signal, not early voting, explains the accuracy. Random pruning loses accuracy at similar compute savings, while hesitation-density pruning preserves SC@32 accuracy even without banking committed answers.

Early voting does a different job, and accuracy at this operating point is the wrong place to look for it. Banking is what keeps the ballot full as the pool narrows, holding the vote at 19.7 of 32 trajectories when only 3.1 are still generating. Its effect on accuracy is neutral here because banking can only change an outcome when pruning would otherwise delete a committed \emph{correct} answer, and at the mild default keep ratio that is rare. It becomes visible once pruning is sharpened: on the hard cells it recovers 3.6pp at a keep ratio of one third, while retaining most of the compute saving (See Appendix~\ref{app:early_voting}). Early voting is therefore what makes the keep ratio safe to expose as a deployment knob, not a source of gain at the default.

Full sweep results, hyperparameter sensitivity, per-model calibration, kernel-size robustness, and the per-difficulty analysis are reported in Appendix~\ref{app:ablations}.

%% file: tables/table_passk_regime.tex
\begin{table}[ht]
\centering
\small
\setlength{\tabcolsep}{4.5pt}
\renewcommand{\arraystretch}{0.95}
\resizebox{\columnwidth}{!}{%
\begin{tabular}{l cc cc}
\toprule
& \multicolumn{2}{c}{\textbf{Hard regime}} & \multicolumn{2}{c}{\textbf{Easy regime}} \\
& \multicolumn{2}{c}{\textit{AIME24+25 (gap 22.3pp)}} & \multicolumn{2}{c}{\textit{AMC+MATH500 (gap 5.3pp)}} \\
\cmidrule(lr){2-3} \cmidrule(lr){4-5}
Method & Acc & FLOP$\downarrow$ & Acc & FLOP$\downarrow$ \\
\midrule
Pass@1 (single rollout)   & 60.7          & --- & 93.8 & --- \\
Pass@32 (oracle ceiling)  & 83.1          & --- & 99.1 & --- \\
\midrule
SC@32 (baseline)          & 75.0          & --- & 96.5 & --- \\
AC                        & 75.2          & 44.9\% & 96.6 & 65.6\% \\
SlimSC                    & 70.3 ($-$4.7) & 64.0\% & 96.1 & 48.3\% \\
\rowcolor{gray!8} \cellcolor{white}
\textbf{FoT@32 (Ours)}    & \textbf{75.0} & \textbf{57.3\%} & \textbf{96.6} & \textbf{41.7\%} \\
\bottomrule
\end{tabular}%
}
\caption{Each method split by pass@$k$ gap. FoT preserves SC@32 accuracy while saving more attention FLOPs in the hard split, where consensus is slow; accuracy is in \%, and FLOP savings are problem-weighted macro averages relative to SC@32.}
\label{tab:passk_regime}
\end{table}

%% file: tables/table3_heldout.tex
\begin{table}[t]
\centering
\resizebox{\columnwidth}{!}{%
\begin{tabular}{l cc r cc}
\toprule
Benchmark & SC@32 & FoT@32 & $\Delta$ (pp) & Token$\downarrow$ & FLOP$\downarrow$ \\
\midrule
\multicolumn{6}{l}{\textit{Qwen3-14B}} \\
AIME24     & 83.3 & 83.3 & $+$0.0 & 32.5\% & 53.6\% \\
AIME25     & 80.0 & 80.0 & $+$0.0 & 38.8\% & 59.6\% \\
AMC23      & 100.0 & 100.0 & $+$0.0 & 18.7\% & 40.1\% \\
MATH500    & 70.4 & 70.6 & $+$0.2 & 12.7\% & 35.6\% \\
\midrule[\cmidrulewidth]
\multicolumn{6}{l}{\textit{Qwen3-32B}} \\
AIME24     & 86.7 & 83.3 & $-$3.3 & 32.3\% & 55.1\% \\
AIME25     & 80.0 & 80.0 & $+$0.0 & 38.3\% & 60.1\% \\
AMC23      & 97.5 & 97.5 & $+$0.0 & 18.7\% & 41.0\% \\
MATH500    & 71.0 & 71.2 & $+$0.2 & 12.2\% & 35.2\% \\
\midrule[\cmidrulewidth]
\multicolumn{6}{l}{\textit{Skywork-OR1-7B}} \\
AIME24     & 80.0 & 80.0 & $+$0.0 & 36.1\% & 56.6\% \\
AIME25     & 60.0 & 60.0 & $+$0.0 & 39.9\% & 60.5\% \\
AMC23      & 95.0 & 95.0 & $+$0.0 & 25.7\% & 53.1\% \\
MATH500    & 70.8 & 70.8 & $+$0.0 & 18.9\% & 48.6\% \\
\midrule[\cmidrulewidth]
\multicolumn{6}{l}{\textit{Phi-4-reasoning}} \\
AIME24     & 50.0 & 56.7 & $+$6.7 & 35.9\% & 58.0\% \\
AIME25     & 36.7 & 53.3 & $+$16.7 & 38.9\% & 61.5\% \\
AMC23      & 55.0 & 67.5 & $+$12.5 & 30.4\% & 54.4\% \\
MATH500    & 46.6 & 47.0 & $+$0.4 & 39.6\% & 57.3\% \\
\midrule
\textbf{Aggregate} & \multicolumn{2}{c}{1600$\to$1615 of 2400} & $+$0.6 & 29.4\% & \textbf{51.9\%} \\
\bottomrule
\end{tabular}%
}
\caption{Cross-model transfer with unchanged FoT configuration: FoT preserves SC@32 accuracy on four held-out models while saving 51.9\% of attention FLOPs on average. Accuracies use exact-match grading, which depresses only the MATH500 absolutes (Appendix~\ref{app:generation}); the deltas are unaffected.}
\label{tab:heldout}
\end{table}

%% file: tables/table_cross_domain.tex
\begin{table}[t]
\centering
\resizebox{\columnwidth}{!}{%
\begin{tabular}{l cc r c ccr}
\toprule
& & & & & \multicolumn{3}{c}{pass@$k$ (LCB only)} \\
\cmidrule(lr){6-8}
Model & SC@32 & FoT@32 & $\Delta$ (pp) & FLOP$\downarrow$ & SC & FoT & $\Delta$ (pp) \\
\midrule
\multicolumn{8}{l}{\textit{GPQA-Diamond (accuracy \%)}} \\
DS-R1-1.5B   & 38.9 & 41.4 & $+$2.5 & 42.2\% & --- & --- & --- \\
DS-R1-7B     & 55.6 & 55.6 & $+$0.0 & 31.8\% & --- & --- & --- \\
OT3-7B       & 51.5 & 53.0 & $+$1.5 & 56.8\% & --- & --- & --- \\
Qwen3-4B     & 65.7 & 67.7 & $+$2.0 & 35.3\% & --- & --- & --- \\
Qwen3-30B    & 72.2 & 72.2 & $+$0.0 & 38.7\% & --- & --- & --- \\
QwQ-32B      & 63.6 & 65.2 & $+$1.5 & 41.4\% & --- & --- & --- \\
\midrule
\textbf{Avg.} & 57.9 & \textbf{59.2} & $+$1.3 & \textbf{41.0\%} & --- & --- & --- \\
\midrule[\cmidrulewidth]
\multicolumn{8}{l}{\textit{LCB-Lite (mean pass rate \% / pass@$k$ \%)}} \\
DS-R1-1.5B   & 15.10 & 16.29 & $+$1.19 & 56.6\% & 35.4 & 34.0 & $-$1.5 \\
DS-R1-7B     & 36.59 & 39.49 & $+$2.90 & 50.3\% & 61.6 & 59.0 & $-$2.6 \\
OT3-7B       & 43.63 & 44.60 & $+$0.97 & 62.2\% & 64.9 & 61.9 & $-$3.0 \\
Qwen3-4B     & 49.23 & 51.78 & $+$2.55 & 53.5\% & 76.9 & 72.4 & $-$4.5 \\
Qwen3-30B    & 69.36 & 69.38 & $+$0.03 & 60.0\% & 83.2 & 81.7 & $-$1.5 \\
QwQ-32B      & 61.30 & 62.15 & $+$0.85 & 57.9\% & 79.1 & 75.4 & $-$3.7 \\
\midrule
\textbf{Avg.} & 45.87 & \textbf{47.28} & $+$1.42 & \textbf{56.8\%} & 66.9 & 64.1 & $-$2.8 \\
\bottomrule
\end{tabular}%
}
\caption{Cross-domain transfer with unchanged FoT configuration. GPQA reports accuracy; LCB reports mean per-rollout pass rate and, in the right block, pass@$k$ (a problem counts as solved if any rollout in the pool---for FoT, any surviving rollout---passes all tests). $\Delta$ is FoT${-}$SC in percentage points.}
\label{tab:cross_domain}
\end{table}

%% file: tables/table5_online.tex
\begin{table}[t]
\centering
\resizebox{\columnwidth}{!}{%
\small
\setlength{\tabcolsep}{5pt}
\renewcommand{\arraystretch}{1.05}
\begin{tabular}{l rrr r}
\toprule
& Pass@1 & SC@32 & \textbf{FoT@32} & vs.\ SC@32 \\
\midrule
Accuracy (\%)                       & 53.3    & 76.7     & \textbf{80.0}     & $+$3.3\,pp \\
Wall time (s)                       & 4{,}145 & 21{,}561 & \textbf{13{,}445} & $-$37.6\% \\
Attn.\ FLOPs ($\times10^{10}$)      & 0.35    & 12.9     & \textbf{5.67}     & $-$56.1\% \\
Cumulative KV (M)                   & 0.37    & 12.88    & \textbf{8.51}     & $-$33.9\% \\
\bottomrule
\end{tabular}
}
\caption{Online deployment on AIME24 with DS-R1-7B ($k{=}32$), measured end-to-end on a single A100. Cumulative KV counts generated-token positions rather than peak memory; pruned and banked rollouts release their slots at each checkpoint.}
\label{tab:online}
\end{table}

%% file: tables/table7_evrp.tex
\begin{table}[t]
\centering

\resizebox{\columnwidth}{!}{%
\setlength{\tabcolsep}{5pt}
\renewcommand{\arraystretch}{0.95}
\begin{tabular}{l cc cc}
\toprule
& \multicolumn{2}{c}{\textbf{w/o Early Voting}} & \multicolumn{2}{c}{\textbf{w/ Early Voting}} \\
\cmidrule(lr){2-3} \cmidrule(lr){4-5}
Pruning Signal & Correct & FLOP$\downarrow$ & Correct & FLOP$\downarrow$ \\
\midrule
None (SC@32)      & \multicolumn{2}{c}{94.36 (baseline)} & \multicolumn{2}{c}{---} \\
Random            & 93.94 ($-$0.42) & 48.0\% & 94.38 ($+$0.01) & 48.4\% \\
Hesitation density & 94.61 ($+$0.25) & 51.7\% & \textbf{94.47 ($+$0.11)} & \textbf{48.7\%} \\
\bottomrule
\end{tabular}%
}

\caption{Pruning signal $\times$ early voting on the 4-benchmark pool (3{,}600 model-problem pairs), with $\Delta$ relative to the SC@32 baseline of 94.36; random rows are means over 10 seeds, with a pooled standard deviation of $\pm$0.14 without early voting and $\pm$0.06 with it. Early voting slightly \emph{lowers} the FLOP saving because banked rollouts are exempt from the pruning quota, so the shrunken active pool ends pruning earlier.}
\label{tab:ablation_evrp}
\end{table}

%% file: sections/5.Conclusion.tex
\section{Discussion}
\label{sec:discussion}

Our findings suggest that predicting per-rollout correctness from hesitation density is fragile, since the same markers appear in both correct and incorrect reasoning and only moderately separate them in aggregate, with $r_{\text{pb}}{=}-0.30$. The reliable signal is distributional: error rate rises monotonically with density, with a decile-level $r=-0.83$, so the high-density tail is disproportionately wrong. This is why FoT works by removing that tail from the answer distribution, pruning the votes least likely to be right and the compute least likely to pay off. It needs no confidence model, reward model, or logit access, and matches SC@32 at roughly half the attention FLOPs, a 28.8\% reduction in full-model FLOPs.

The key reason this replacement transfers is that FoT does not rely on a model-specific threshold for how often a word such as \textit{wait} or \textit{actually} should appear. Absolute marker frequency varies across LRMs, but failed long-form reasoning often exposes a shared lexical pattern: repeated revision, backtracking, and non-commitment. FoT uses this pattern comparatively, ranking active rollouts within the same model, problem, and checkpoint. As a result, the signal can survive architectural changes and domain shifts: a Qwen-family model, Phi-4-reasoning, a graduate-level science question, and a code-generation task need not use identical wording for FoT to identify the unusually unproductive tail of the current pool.

This also clarifies why FoT differs from the efficient-SC baselines. Adaptive Consistency is strongest when the answer distribution stabilizes quickly, and Slim-SC is strongest when similar trajectories are genuinely redundant. Those are precisely the cases where multiple sampling contributes less. In the harder regime, consensus may be slow and superficially similar derivations may still carry useful voting evidence. FoT is designed for this regime: it preserves the pool-level diversity that makes SC effective, while pruning within-trajectory waste before it dominates the compute budget. This is why its advantage is most persuasive on hard cells, held-out models, and domain shifts, where robustness matters more than early consensus.

%% file: sections/6.Limitations.tex
\section{Limitations}

Our evaluation is centered on reasoning tasks with extractable final answers. We include cross-domain tests on GPQA-Diamond and LCB-Lite, but these still have relatively well-defined answer or execution-based evaluation. Open-ended generation, multi-turn interaction, and tasks where correctness cannot be reduced to answer extraction remain outside the scope of this study.

Our calibration pool is also unbalanced by construction: MATH500 supplies 3{,}000 of the 3{,}600 model-problem pairs, so any statistic pooled over the whole suite is weighted toward easy problems on which pruning has little to remove and little to risk. We report the hard split separately throughout for this reason, and the hard-split numbers are the ones to read when judging the mechanism rather than the aggregate. A pool built with a different difficulty mix would move the pooled figures, though it should not affect the within-pool relative comparison that FoT actually uses.

FoT also relies on a commitment detector for early voting. In our math setting this is the \texttt{\textbackslash boxed\{\}} convention, while other domains require task-specific answer extraction or prompting. The pruning signal itself is independent of the final-answer format, but the vote bank is only as reliable as the commitment detector.

Finally, FoT uses a fixed checkpoint schedule and an English hesitation-marker kernel. The relative-density rule reduces sensitivity to model-level verbosity, but models reasoning in other languages or with substantially different deliberation styles may require revalidating the marker set. Our FLOP accounting assumes standard full attention; under efficient-attention mechanisms the saving scales down toward the token-reduction floor (from 48.7\% to 23.0\% in the linear limit), which we quantify across window sizes in Appendix~\ref{app:efficient_attention}. Pruning still releases KV cache and batch slots regardless of the attention regime. Adaptive checkpointing based on per-problem difficulty is another natural extension that could further improve the efficiency--accuracy tradeoff.

%% file: sections/x.Appx.tex
\section{Motivating Analysis}
\label{app:motivating}

\input{tables/app3_tokens}

\subsection{Hesitation-Marker Validation}
\label{app:tokens}

Table~\ref{tab:tokens} shows the 21 hesitation markers used by FoT. We select markers whose density has a significant negative \emph{benchmark-weighted} point-biserial correlation with rollout correctness (weighted $r<0$, weighted $p<0.05$); weighting each benchmark equally prevents MATH500's 96{,}000 rollouts from dominating the pooled statistic, and all 21 markers satisfy the criterion. The resulting kernel spans common revision markers such as \texttt{Wait,} and \texttt{actually}, as well as rarer but highly skewed phrases such as \texttt{Let me reconsider}.

We exclude markers that correlate positively with correctness, such as \texttt{mistake}, \texttt{Oh}, and \texttt{Hmm}, because they more often indicate productive self-correction. The aggregate density of the 21 retained markers yields $r_{\text{pb}}=-0.30$, making the composite signal stronger than most individual markers.

Table~\ref{tab:density_decile} gives the decile-level diagnostic behind Observation~3. Accuracy declines from 75.7\% in the lowest-density decile to 50.7\% in the highest-density decile. The signal is not intended to identify the single correct rollout; rather, it separates the unproductive high-density tail that FoT should prune.

\paragraph{Density and eventual length.} Decomposing the two signals per rollout on AIME24, final rollout length correlates with correctness at $-0.60$ to $-0.76$ across models and hesitation density at the 8K checkpoint at $-0.05$ to $-0.38$, while the partial correlation of density with correctness given final length is small and model-dependent ($-0.24$ to $+0.10$). Density instead forecasts eventual length from the visible prefix ($r{=}{+}0.13$ to ${+}0.41$), which is the sense in which it is a leading indicator of runaway generation (See Observation~3). It therefore carries little information beyond eventual length; its value is that it is available online. An oracle that prunes by \emph{final} length reaches the same accuracy as FoT (72.9 on the exact-match lineage of Table~\ref{tab:ablation_evrp}) at a 61.8\% attention saving against FoT's 50.7\%, but a rollout's final length is unknown until it has been paid for, and at a fixed token-count checkpoint all active rollouts have the same length, so ``prune the longest'' is not even defined online.

\input{tables/app_decile_density}

\input{tables/app_model_density}

\subsection{Marker-Selection Validation}
\label{app:crossval}

A potential concern is that the marker list may overfit to the lexical styles of the models and benchmarks used for calibration. We address this primarily through the held-out model and cross-domain experiments in Section~\ref{sec:cross_model} and Section~\ref{sec:cross_domain}, where the same 21-marker kernel is applied without retuning.

Table~\ref{tab:model_density} reports the absolute marker density by model. The scale varies substantially, but FoT does not apply a fixed density threshold across models; it ranks rollouts within the same active pool at the same checkpoint. This relative rule is why a shared marker kernel can transfer across models with different verbosity levels.

\input{tables/app_phi4_pathology}

\paragraph{Phi-4-reasoning diagnostic.} Table~\ref{tab:phi4_pathology} reports the rollout profile behind the Phi-4 cross-model result in Section~\ref{sec:cross_model}. Phi-4 has both a large pass@32--SC@32 gap and an unusually high number of no-answer rollouts, making it a useful diagnostic for whether FoT can remove non-committing trajectories without retuning.

\input{tables/app_no_answer_reduction}

\paragraph{No-answer tail diagnostic.} Table~\ref{tab:no_answer_reduction} reports the checkpoint-level analysis used to interpret Figure~\ref{fig:typology}. FoT does not identify every failed rollout independently; instead, it progressively removes the non-committing tail from the active pool, with the largest reductions in the cells where no-answer behavior is most acute.

\input{tables/app1_gap}

\subsection{Pass@1 vs.\ Pass@32 Performance Gap}
\label{app:passk_gap}

Table~\ref{tab:passk_gap} reports the per-rollout accuracy, oracle accuracy, and SC@32 majority-vote accuracy for all 6 models across 4 benchmarks. The average pass@1--pass@32 gap is 14.4 percentage points, confirming that individual rollouts are highly stochastic and that correct answers often appear somewhere in the 32-rollout pool. The gap is largest for smaller models on harder benchmarks and smallest for stronger models near ceiling.

\section{Additional Experiments}
\label{app:experiments}

This section collects secondary diagnostics behind the main results: per-model behavior, the cases where FoT and SC@32 disagree, benchmark and per-problem difficulty effects, and a unified-basis comparison with sample-axis baselines.

\subsection{Per-Model Accuracy Breakdown}
\label{app:per_model}

\begin{figure*}[h]
\centering
 \includegraphics[width=\textwidth]{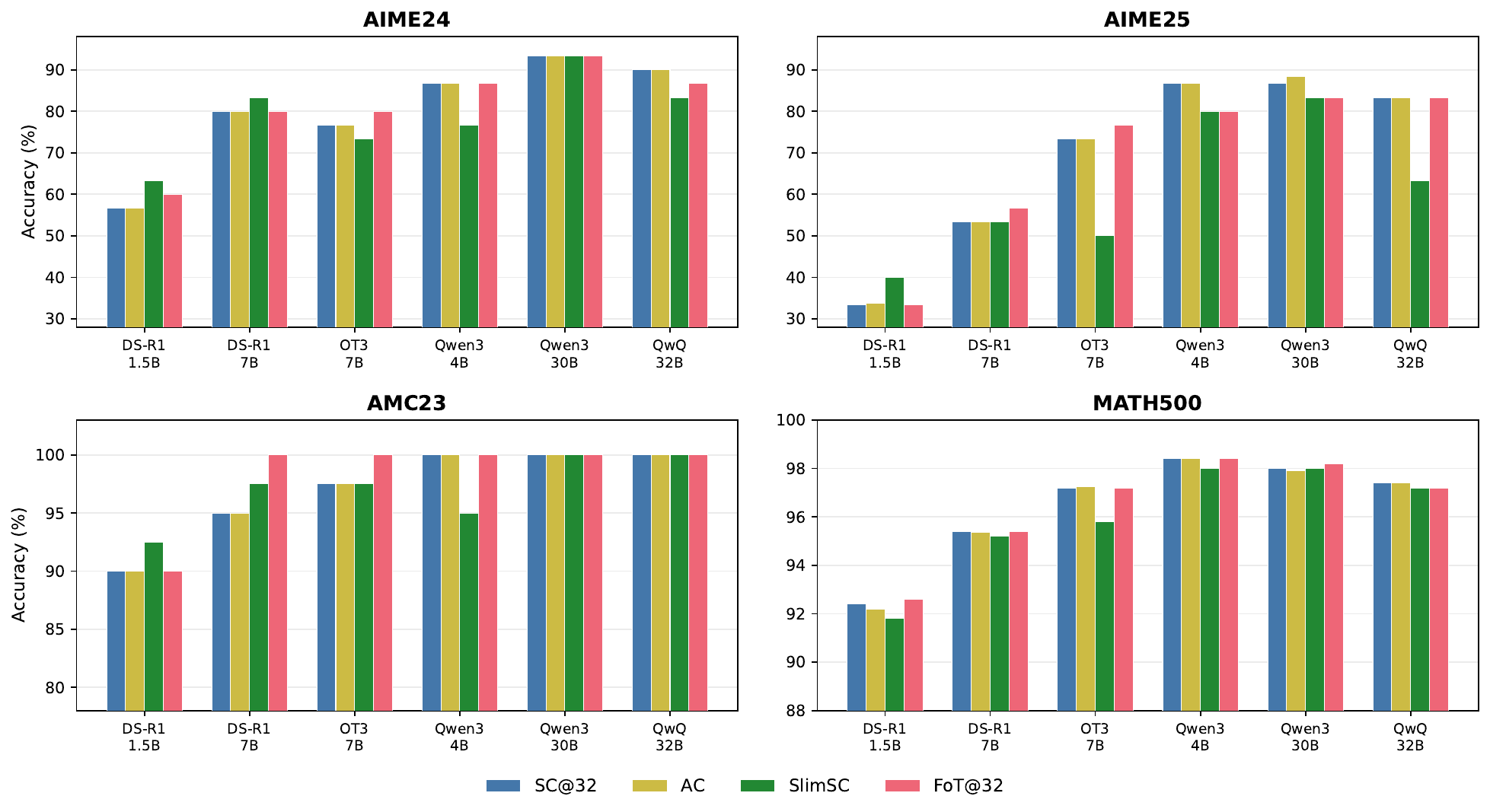}
\caption{Per-model accuracy breakdown across the 4-benchmark math suite.}
\label{fig:appx_1}
\end{figure*}

Figure~\ref{fig:appx_1} breaks down accuracy by model on the 4-benchmark math suite. The pattern mirrors the main table: FoT is most helpful when the base model still has substantial sampling variance, while near-ceiling models leave less room for improvement. SlimSC is less stable on the harder AIME split, consistent with the main-text observation that similarity pruning can remove useful vote diversity.

\subsection{Discordant Instance Analysis}
\label{app:discordant}

\begin{figure}[h]
\centering
 \includegraphics[width=\columnwidth]{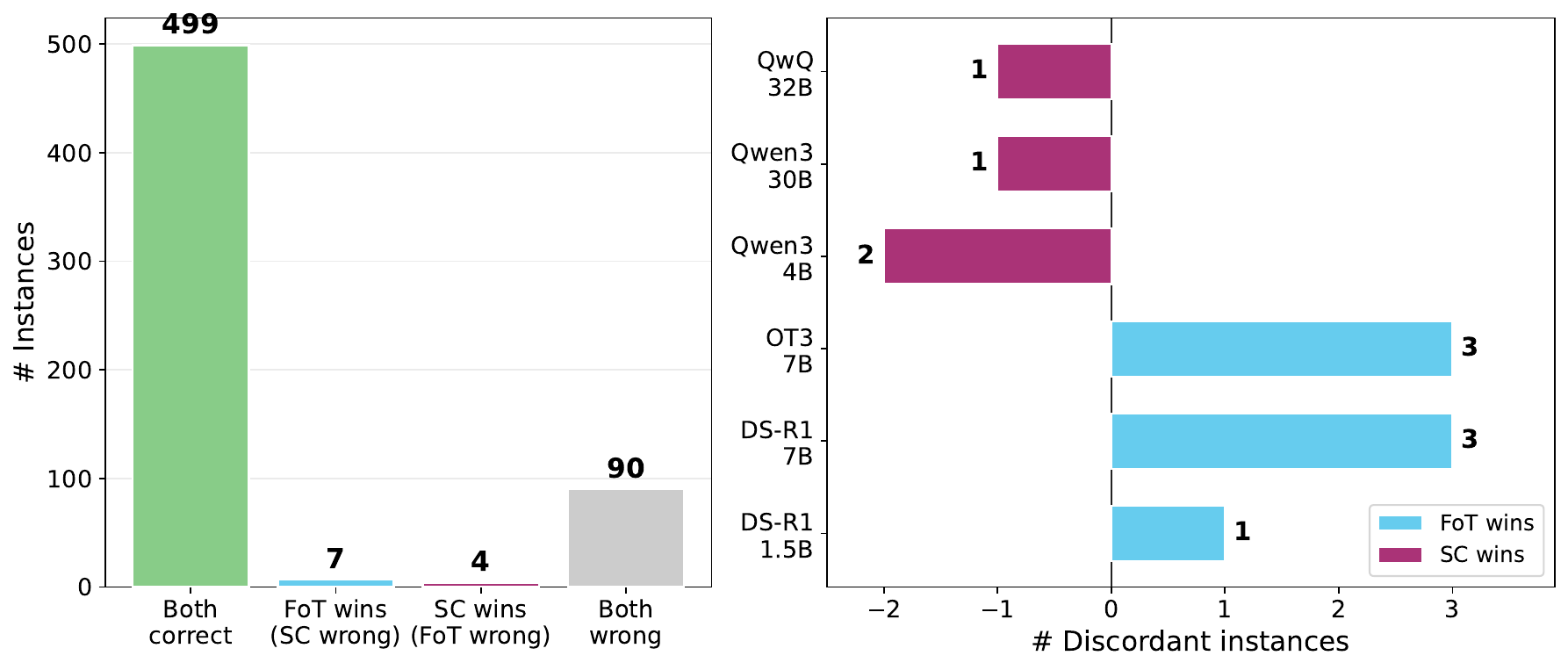}
\caption{Discordant cases on the AIME/AMC subset. Left: SC@32--FoT@32 concordance. Right: FoT wins and losses by model.}
\label{fig:appx_4}
\end{figure}

Figure~\ref{fig:appx_4} examines cases where FoT and SC@32 produce different correctness outcomes on the AIME/AMC subset. FoT wins more often than it loses, and the losses cluster on harder problems with low pass rates. This supports the main-text interpretation that FoT usually preserves the SC answer distribution, while occasionally changing the final vote after removing non-committing or repetitive trajectories.

\subsection{Vote-Concentration Diagnostic}
\label{app:vote_concentration}

\begin{figure}[h]
\centering
 \includegraphics[width=\columnwidth]{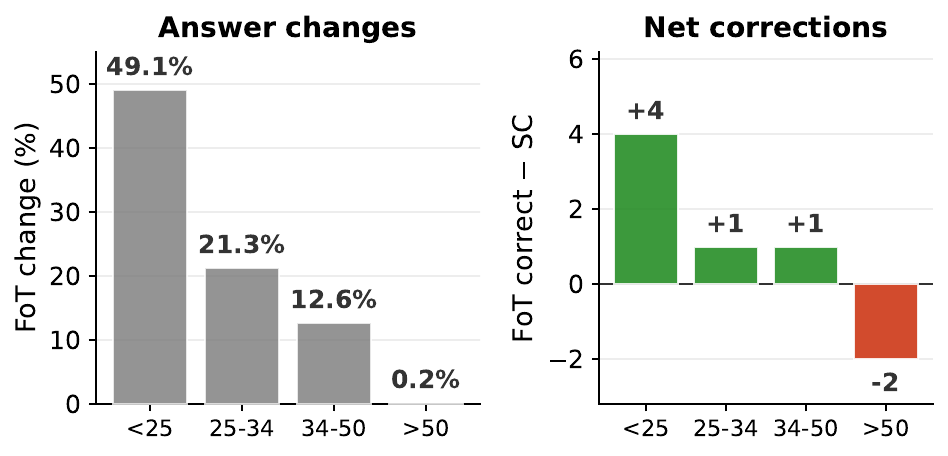}
\caption{Vote-concentration diagnostic on the 4-benchmark pool. Bins are SC@32 plurality-share ranges. Left: how often FoT changes the SC answer. Right: FoT correct minus SC correct within each bin.}
\label{fig:vote_concentration}
\end{figure}

Figure~\ref{fig:vote_concentration} asks when FoT changes the final SC decision. We bin each model-problem pair by the fraction of rollouts assigned to the SC@32 plurality answer. Four pairs with no extractable SC answer are excluded from this diagnostic, leaving 3{,}596 pairs.

The pattern is consistent with FoT's intended role. When SC@32 has a clear majority, FoT almost always preserves it: in the $>50\%$ bin, the final answer changes in only 0.2\% of cases. In contrast, low-consensus pools are both more diverse and less reliable: the $<25\%$ bin averages 15.2 distinct answers, and FoT changes the final answer in 49.1\% of cases. This is also where most net corrections arise ($+4$). The effect is modest, but it explains why FoT can slightly improve aggregate accuracy while primarily serving as a compute-reduction method: it leaves strong SC trends intact and acts mainly on the divergent tail.

\subsection{Difficulty Adaptation}
\label{sec:cross_dataset}

MATH500 \citep{lightman2023letsverifystepstep} sits at the easier end of our math suite. Table~\ref{tab:main} shows that FoT maintains accuracy parity on MATH500 while saving less compute than on AIME/AMC. This is expected: easier problems produce shorter rollouts with less hesitation tail to prune, so many trajectories commit early instead of entering the pruning regime. The result clarifies that FoT's savings track late-stage reasoning waste rather than applying a fixed reduction uniformly across benchmarks.

\subsection{Difficulty-Stratified Accuracy}
\label{app:stratification}

Table~\ref{tab:stratification} bins every model-problem pair by per-problem pass@1, the fraction of its 32 rollouts that are correct. FoT sits 4.8--5.8pp above SC@32 in the low-pass@1 bins ($0{<}p_1{\leq}0.25$), where the pass@1--pass@32 gap is largest and the vote is least concentrated. The $p_1{=}0$ bin is unreachable by any prune-only method, since no correct rollout exists in the pool to preserve, and the saturated $p_1{>}0.5$ bin is at ceiling. This is the same picture as the vote-concentration diagnostic of Section~\ref{app:vote_concentration}, resolved by difficulty rather than by plurality share.

\input{tables/table_stratification}

\subsection{Sample-Axis Baselines on a Unified FLOP Basis}
\label{app:baselines}

FoT prunes along the token axis, whereas adaptive-consistency methods act on the sample axis by deciding how many rollouts to draw. Table~\ref{tab:baselines} places both on one basis: \texttt{math\_verify} grading and full-model (attention, FFN, and projection) FLOP savings against SC@32. At equal pooled accuracy DSC saves more than FoT overall (69.4\% vs.\ 28.8\%), and our offline Certaindex proxy also holds SC accuracy at an 11.5--13.2\% saving, though the proxy is only indicative: it can fire later than the original online probe, understating savings, while paying no probe cost, overstating them.

The two axes move in opposite directions with difficulty. DSC's saving falls from 75.0\% on MATH500 to 54.4\% on AIME24/25 as consensus slows, while FoT's rises from 21.9\% to 44.0\% because hard problems carry longer unproductive tails. Because the axes are independent, the methods compose: FoT+DSC reaches 80.9\% (all) and 75.7\% (hard) full-model savings, at accuracy costs of $-0.4$ and $-1.4$pp. Per-query latency and KV-cache relief remain specific to FoT, as a sample-axis method cannot shorten a rollout it has already drawn.

\input{tables/table_baselines}

\section{Additional Ablations}
\label{app:ablations}

\begin{figure*}[h]
\centering
 \includegraphics[width=\textwidth]{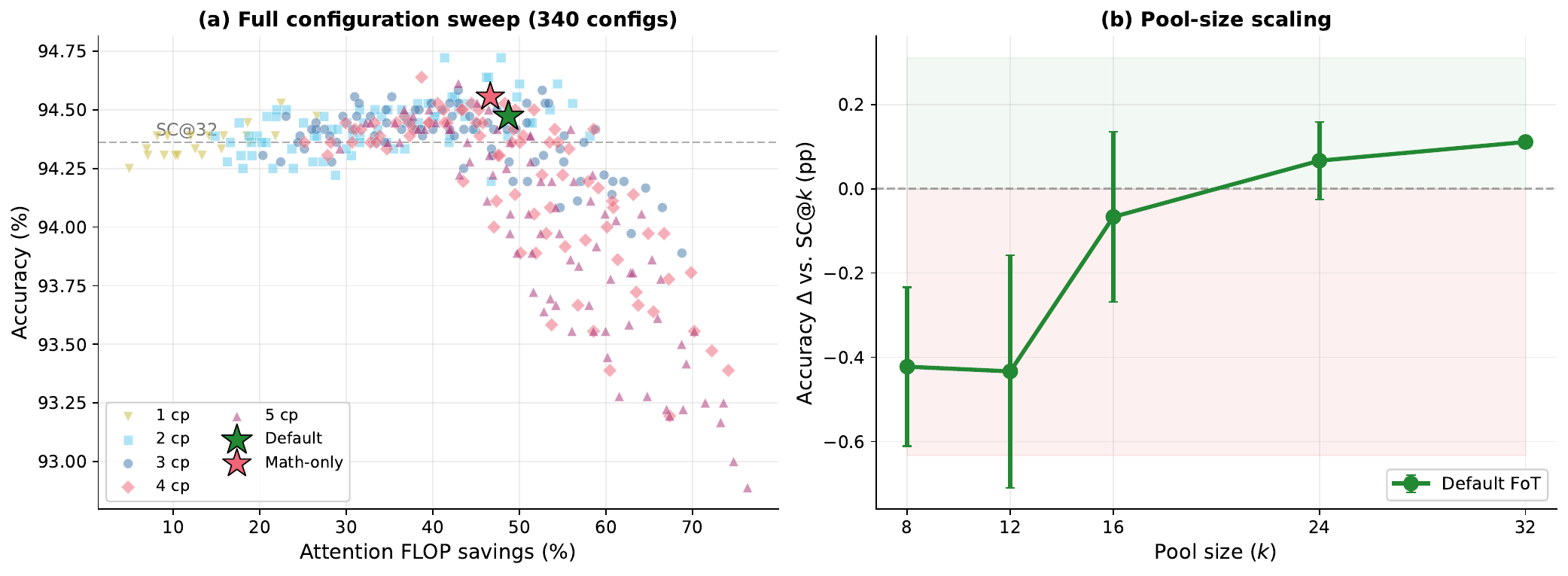}
\caption{Hyperparameter sensitivity analysis. \textbf{(a)} Full 340-configuration sweep on the 4-benchmark calibration pool. Stars mark the default joint-Pareto config and the math-only in-domain maximizer. \textbf{(b)} Pool-size scaling for the default config.}
\label{fig:sensitivity}
\end{figure*}

\subsection{Hyperparameter Sensitivity}
\label{app:sensitivity}

To validate our hyperparameter choices, we conduct a comprehensive sweep over 340 configurations spanning checkpoint count (1--5), checkpoint placement, and keep ratio ($\rho \in \{0.50, 0.60, 0.67, 0.75, 0.85\}$) on the unified 4-benchmark calibration pool (6 models $\times$ 600 problems $\times$ 32 rollouts $=$ 115{,}200 rollouts, with inverse-frequency stratified weighting so MATH500's larger problem count does not dominate the smaller benchmarks). Figure~\ref{fig:sensitivity}(a) plots all configurations on accuracy vs.\ FLOP savings axes, color-coded by checkpoint count.

Two natural Pareto candidates emerge. The \emph{math-only} winner ($\{4\text{K}, 14\text{K}\}$, $\rho{=}0.60$) maximizes accuracy on the math calibration pool ($\Delta{=}{+}13$ / 3{,}600 problems, ${+}0.36$pp) at a 47.9\% FLOP saving. The \emph{joint} winner ($\{6\text{K}, 10\text{K}, 14\text{K}\}$, $\rho{=}0.67$; green star), which we use throughout the paper, gives up 9 of those math problems while saving marginally more compute (48.7\%), in exchange for substantially larger gains on out-of-distribution evaluations: ${+}15$ problems on GPQA-Diamond (${+}1.26$pp on 1{,}188 model-problem pairs) and ${+}14$ problems on the cross-architecture Phi-4-reasoning evaluation (${+}2.33$pp on 600 model-problem pairs spanning the same four math benchmarks).

\paragraph{Why not the math-only winner?} The math-only schedule is the best choice if the sole objective is maximizing the four-benchmark math aggregate. We instead use the joint configuration because the paper's claim is model- and domain-level transfer: it gives up 9 math problems but recovers larger gains on the held-out Phi-4 and GPQA evaluations. It ranks 71st of the 340 configurations on the math pool, so 58 configurations beat it in domain; this makes the default a deliberately conservative operating point rather than the in-domain maximizer. The out-of-distribution figures quoted above are graded by exact match, the lineage of the held-out evaluations in Table~\ref{tab:heldout}, while the math-pool figures use \texttt{math\_verify}. On GPQA that choice is nearly immaterial: the two graders assign different SC@32 outcomes on 8 of the 1{,}188 model-problem pairs, every one of them a single model emitting \texttt{\textbackslash text\{X\}} in place of the bare option letter.

\paragraph{Pool-size behavior.} Figure~\ref{fig:sensitivity}(b) shows the expected tradeoff for the default schedule: pruning is too aggressive when the initial pool is small, but the gap closes as $k$ grows and becomes slightly positive at $k{\geq}24$. This supports using FoT primarily as a replacement for large-pool SC rather than as a small-$k$ method.

\subsection{Checkpoint and Threshold Sensitivity}
\label{sec:ablation_sweep}

We sweep four checkpoint schedules (Early: $\{4\text{K}, 8\text{K}, 12\text{K}\}$, Mid: $\{6\text{K}, 10\text{K}, 14\text{K}\}$, Math-only: $\{6\text{K}, 11\text{K}, 16\text{K}\}$, Late: $\{8\text{K}, 12\text{K}, 16\text{K}\}$) crossed with three keep ratios ($\rho \in \{0.50, 0.67, 0.75\}$) on the unified 4-benchmark pool (3{,}600 model-problem pairs per cell). Table~\ref{tab:ablation} shows the results.

\input{tables/table6_ablation}

Two patterns emerge. First, the Early schedule is consistently the worst, hurting accuracy at every keep ratio ($\Delta \leq {-}2$, down to ${-}15$ at $\rho{=}0.50$): early checkpoints prune before the hesitation marker signal has accumulated sufficiently, removing rollouts that would have converged correctly given more tokens. Second, the Mid, Math-only, and Late schedules all produce viable configurations ($\Delta \geq 0$) at $\rho \in \{0.67, 0.75\}$, with the math-only winner reaching $\Delta{=}{+}9$ on math but giving up the larger out-of-distribution gains discussed in Section~\ref{app:sensitivity}. The universal Mid schedule with $\rho{=}0.67$, used throughout the paper, is the joint Pareto choice balancing math accuracy ($\Delta{=}{+}4$) against the larger OOD gains.

\subsection{Per-Model Calibration}
\label{sec:ablation_permodel}

All main experiments use a single universal config (Mid schedule $\{6\text{K}, 10\text{K}, 14\text{K}\}$, $\rho{=}0.67$). We ask: does per-model hyperparameter tuning improve results? For each model, we sweep 12 configurations (4 schedules $\times$ 3 keep ratios) on that model's 600 problems across the four math benchmarks independently.

\input{tables/table8_permodel}

Table~\ref{tab:permodel} shows that per-model calibration yields a small aggregate improvement: $\Delta{=}{+}17$ vs.\ universal ${+}4$ on 3{,}600 model-problem pairs (a marginal ${+}13$ problems, ${+}0.36$pp). The bulk of this gain concentrates on the weakest model in the suite (DS-R1-1.5B picks Math-only $\rho{=}0.50$ for $\Delta{=}{+}10$ vs.\ universal ${+}3$); the remaining five models gain $\leq 2$ problems each from per-model tuning. Three of six models select the Math-only schedule $\{6\text{K}, 11\text{K}, 16\text{K}\}$, matching the in-domain tradeoff described above, but the preferred schedules are not stable enough to justify model-specific tuning. We therefore keep the universal config as the default.

\input{tables/app2_transfer}

\subsection{Per-Model Config Transfer}
\label{app:transfer}

To test whether per-model configs transfer, we apply each of the five distinct optimal configs from Section~\ref{sec:ablation_permodel} to all 6 models. Table~\ref{tab:transfer} shows that the math-only Pareto winner, Math-only $\{6\text{K}, 11\text{K}, 16\text{K}\}$ with $\rho{=}0.67$, achieves the highest cross-model math aggregate ($\Delta{=}{+}9$ on 3{,}600), but this is the same in-domain tradeoff described in Section~\ref{app:sensitivity}. Our universal Mid config ($\rho{=}0.67$, $\Delta{=}{+}4$ on math) is the joint Pareto choice that balances math performance against out-of-distribution generalization. No per-model-tuned configuration dominates the universal config across both math and OOD evaluations simultaneously.

\input{tables/app_kernel_sweep}

\subsection{Kernel-Size Robustness}
\label{app:kernel_sweep}

We sweep the hesitation-marker kernel from 1 to 21 markers, adding markers in descending order of $|r_{\text{pb}}|$ (Table~\ref{tab:tokens}). All other hyperparameters are held at the paper default ($\{6\text{K}, 10\text{K}, 14\text{K}\}$, $\rho{=}0.67$, hesitation marker density). Table~\ref{tab:kernel_sweep} reports the full sweep on the unified 4-benchmark pool (3{,}600 model-problem pairs).

Two findings. First, accuracy is essentially flat across kernel sizes: $\Delta$ ranges from $-3$ to $+6$ problems on 3{,}600, with all 8 settings within 0.17pp of one another and well within single-rollout sampling noise. Even the top-1 kernel (\texttt{perhaps} alone) preserves SC@32 accuracy to within 3 problems. Second, FLOP saving is constant to two significant figures (50.59--50.76\%), confirming that the pruning behavior is driven by aggregate hesitation density rather than the identity of any individual marker. The three highest-$|r|$ markers (\texttt{perhaps}, \texttt{Wait,}, \texttt{Wait}) match the full 21-marker kernel within sampling noise ($\Delta{=}{+}5$ vs.\ ${+}4$).

This robustness is itself the substantive finding: the lexical signal is concentrated in a small number of canonical English hesitation markers, and the method does not depend on the specific composition of the kernel. We retain the 21-token kernel as the calibrated default for stability, but smaller kernels (down to top-3) are an equivalent operating point.

\subsection{Early Voting Analysis}
\label{app:early_voting}

\begin{figure*}[h]
\centering
 \includegraphics[width=\textwidth]{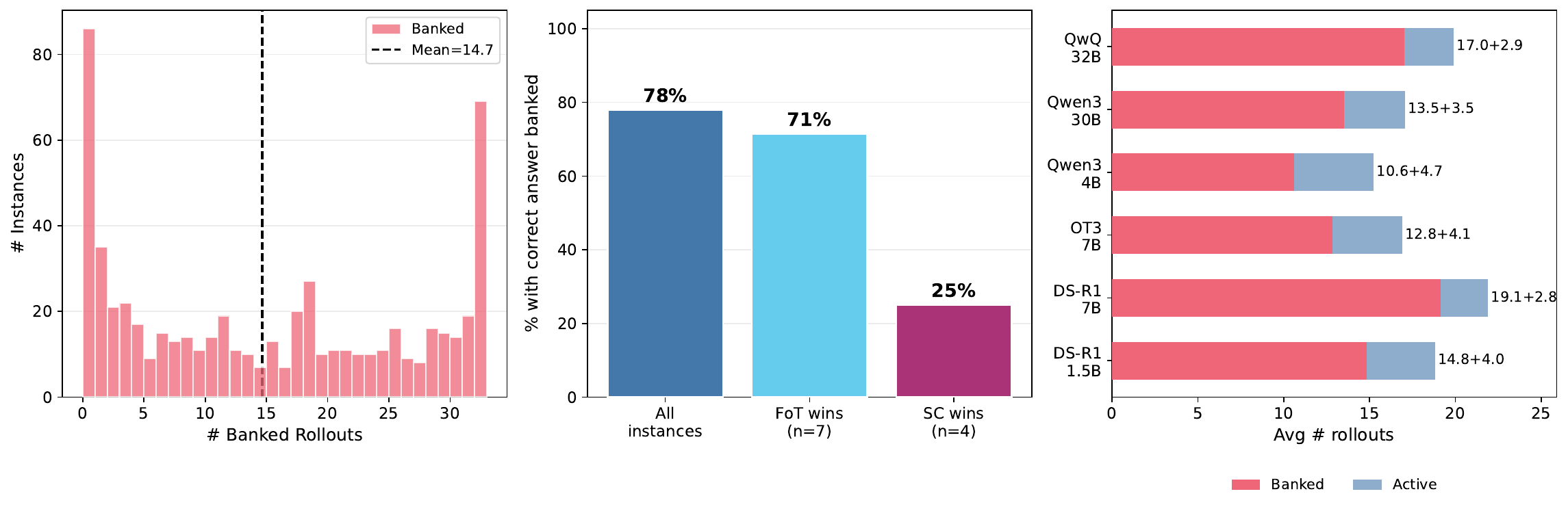}
\caption{Early voting diagnostic on the AIME/AMC subset. Left: banked rollout count per problem. Center: fraction with a correct answer banked. Right: average banked and active rollouts per model.}
\label{fig:appx_2}
\end{figure*}

Figure~\ref{fig:appx_2} analyzes the early voting mechanism on the AIME/AMC subset under the current FoT configuration. On average, 14.7 of 32 rollouts are banked before pruning begins. The bank contains a correct answer in 78\% of all instances, (71\% restricted to FoT wins, 25\% on SC wins). Early commitment is therefore a strong correctness signal, yet the 4-benchmark ablation in Table~\ref{tab:ablation_evrp} shows the aggregate gain coming primarily from hesitation-density pruning. These two facts are consistent, and the reconciliation defines early voting's role.

The neutrality in Table~\ref{tab:ablation_evrp} is structural rather than evidence of a redundant mechanism. The pooled ablation is dominated by MATH500 (3{,}000 of 3{,}600 model-problem pairs), where problems are easy, commitment is fast, and the mild default keep ratio $\rho{=}0.67$ almost never endangers a committed answer; banking answers that pruning would not have touched changes nothing. Early voting's contribution concentrates exactly where this condition fails: hard problems under aggressive pruning. Table~\ref{tab:ev_keepratio} sweeps the keep ratio on the hard cells (AIME24/25, 6 models, $n{=}360$; SC@32 $=$ 75.0):

\begin{table}[h]
\centering
\resizebox{\columnwidth}{!}{%
\begin{tabular}{c cc c cc}
\toprule
& \multicolumn{2}{c}{Accuracy} & & \multicolumn{2}{c}{FLOP$\downarrow$} \\
\cmidrule(lr){2-3} \cmidrule(lr){5-6}
keep $\rho$ & w/ EV & w/o EV & EV $\Delta$ & w/ EV & w/o EV \\
\midrule
0.67 (default)$^{*}$ & 75.0 & 75.0 & $+$0.0 & 57.3\% & 59.2\% \\
0.50 & 73.9 & 73.3 & $+$0.6 & 72.0\% & 73.3\% \\
0.33 & 71.1 & 67.5 & $\mathbf{+3.6}$ & 80.5\% & 81.7\% \\
0.25 & 70.8 & 67.2 & $\mathbf{+3.6}$ & 81.7\% & 82.8\% \\
\bottomrule
\end{tabular}}
\caption{Early voting under keep-ratio sweep on the hard cells (attention-FLOP convention of Table~\ref{tab:main}). At the default $\rho$ early voting is neutral; at aggressive $\rho$ it recovers committed correct answers that pruning would otherwise delete. $^{*}$The default row quotes the main-result lineage (Table~\ref{tab:passk_regime}); the sweep recomputation differs by one problem on one model due to re-tokenization, within single-problem resolution.}
\label{tab:ev_keepratio}
\end{table}

At mild pruning almost no committed answer is ever at risk, so early voting is neutral. As pruning sharpens on hard problems it begins deleting committed \emph{correct} answers, and early voting recovers them: without it, pushing to $\rho{=}0.33$ for ${\sim}82\%$ attention-FLOP savings costs 7.5pp (75.0${\to}$67.5); with it, the same aggressive setting holds 71.1 at 80.5\% savings. Early voting is thus a cheap, always-on safety margin that caps the cost of pruning errors, operationalizing Observation~1: once a rollout has produced \texttt{\textbackslash boxed\{\}} its outcome is resolved, so ranking it with a predictor of future spiraling would apply the signal outside its domain of validity. This safeguard is what makes the keep ratio safe to expose as a deployment knob rather than a fixed constant.

\section{Implementation Details}
\label{app:implementation}

\subsection{Generation Details}
\label{app:generation}

All rollouts are generated using SGLang \citep{zheng2024sglangefficientexecutionstructured} with the following parameters: temperature $T{=}0.6$, top-$p{=}0.95$, top-$k{=}30$, and a maximum generation budget of 32,768 tokens. Each model uses a rolling random seed starting from 42 (i.e., seed $= 42 + \text{rep\_id}$) to ensure reproducibility while maintaining independence across rollouts.

Offline rollouts are collected across multiple GPU configurations depending on model size. For the online deployment experiment in Section~\ref{sec:online_exp}, all three conditions (Pass@1, SC@32, FoT@32) are run on a single isolated A100 GPU with a fixed random seed of 2000, ensuring that the same rollout pool is generated across conditions for a controlled comparison.

\paragraph{Offline simulation and answer handling.} Five implementation details affect exact reproduction. (1) The offline experiments approximate token-count checkpoints by per-rollout linear interpolation over characters; this can move a hesitation marker or a \texttt{\textbackslash boxed\{\}} commitment across a checkpoint boundary relative to a true tokenizer cut, whereas online deployment (\S\ref{sec:online_exp}) cuts on real token counts. (2) Answer extraction (\texttt{extract\_boxed}) falls back to multiple-choice-letter and ``Final Answer:'' patterns when no \texttt{\textbackslash boxed\{\}} is present; these fallbacks can also fire on math outputs. (3) Overlapping markers double-count by design: text matching \texttt{Wait,} also increments the count for \texttt{Wait}. (4) Plurality ties in the final vote are broken by insertion order. (5) On MATH500, 38\% of gold answers are non-integer LaTeX expressions, which is why only MATH500 absolute accuracies were depressed under the earlier exact-string-match grader while the integer-answer AIME/AMC benchmarks were unaffected; all main-result accuracies in this version are graded with \texttt{math\_verify}; appendix analyses that retain the earlier exact-match grading are labeled where they appear.

\paragraph{Checkpoint synchronization overhead.} FoT requires pausing all active rollouts at each token-count checkpoint to evaluate the pruning criterion. In batch-mode serving frameworks such as SGLang, this synchronization is natively supported: all rollouts in a batch share the same generation schedule, so pausing at a fixed token count introduces no idle time. In continuous-batching frameworks (e.g., vLLM), rollouts that reach a checkpoint before others may incur brief idle time while the batch synchronizes, a minor scheduling cost. Our online wall-time measurements in Section~\ref{sec:online_exp} confirm that this overhead is negligible in practice: the pruning decision itself is a string count operation requiring no model inference.

\paragraph{KV-cache footprint.} In the online deployment run, FoT reduces the cumulative generated-token footprint from 12.88M under SC@32 to 8.51M under FoT@32, a 33.9\% reduction. Since each generated token occupies one KV-cache position per layer, this is the memory-side counterpart of the wall-time and generation-FLOP savings reported in Table~\ref{tab:online}.

\subsection{Efficient-Attention Scaling}
\label{app:efficient_attention}
Our headline FLOP accounting (Section~\ref{sec:prelim}) assumes standard full attention, where per-rollout cost scales as $s^2$ and late tokens dominate, which is precisely the cost FoT removes by terminating spiraling rollouts early. Efficient attention changes this accounting: under sliding-window attention with window $w$, per-token cost saturates once the position exceeds $w$, so per-rollout cost grows linearly ($\approx w\cdot s$) rather than quadratically, and sparse or linear variants behave similarly. FoT's saving then tracks its token reduction rather than the larger quadratic figure. Table~\ref{tab:efficient_attention} applies FoT's termination pattern under each regime on the full 4-benchmark pool. The full-attention saving (48.7\%, pooled over all rollouts as in Table~\ref{tab:ablation_evrp}) degrades smoothly toward the token-reduction floor (23.0\%) as the window narrows; realistic sliding-window models ($w \approx 4$--$8$K) retain roughly 63--79\% of it. The accuracy result is unaffected, since it depends only on which rollouts survive pruning, and the KV-cache release at each checkpoint is likewise independent of the attention regime. Three headline savings therefore coexist in this paper as related but distinct decompositions of the same pruning pattern: the pooled full-quadratic attention saving (48.7\%), the linear-attention token-reduction floor (23.0\%), and the full-model FLOP saving that includes non-attention compute (28.8\%; Table~\ref{tab:baselines}).

\begin{table}[t]
\centering
\small
\begin{tabular}{l r}
\toprule
Attention regime & FLOP saving \\
\midrule
Full quadratic (headline)  & 48.7\% \\
Sliding window, $w{=}8$K    & 38.6\% \\
Sliding window, $w{=}4$K    & 30.9\% \\
Sliding window, $w{=}2$K    & 26.8\% \\
Sliding window, $w{=}1$K    & 24.8\% \\
Linear / sparse top-$k$     & 23.0\% \\
\bottomrule
\end{tabular}
\caption{FoT attention-FLOP saving relative to SC@32 under different attention regimes, on the 4-benchmark pool with the universal configuration. The saving interpolates between the full-attention headline and the token-reduction floor as the effective attention window narrows.}
\label{tab:efficient_attention}
\end{table}

\subsection{Adaptive Consistency Baseline}
\label{app:ac_config}

We simulate Adaptive Consistency (AC; \citealp{aggarwal-etal-2023-lets}) on the same pre-generated rollout pool used by all other methods.\footnote{\url{https://github.com/Pranjal2041/AdaptiveConsistency}} AC is a sequential early-stopping method: rollouts are consumed one at a time, and at each step a Bayesian stopping criterion determines whether the current majority answer is sufficiently confident. We use the Beta stopping criterion with confidence threshold $\tau{=}0.95$ (the paper's default), which stops when posterior confidence in the current majority answer exceeds $\tau$.

Because AC assumes sequential generation while our rollouts are generated in parallel, we simulate sequentiality by processing rollouts in index order. To account for ordering effects, we average results over 10 random permutations of the rollout sequence per problem. AC's compute savings are measured as the fraction of rollouts \emph{not consumed} before stopping; FLOP savings are computed identically to other methods (sum of $s_i^2$ over consumed rollouts only).

\subsection{Slim-SC Baseline Configuration}
\label{app:slimsc_config}

Our Slim-SC baseline is a faithful reproduction of \citet{hong2025slimscthoughtpruningefficient}, using the configuration from their official repository.\footnote{\url{https://github.com/hyscale-lab/slimsc}} We use the diversity pruning strategy with cosine similarity threshold $\tau{=}0.9$, segment-level embeddings via \texttt{sentence-transformers/all-mpnet-base-v2}, and a warm-up of 20 thoughts before pruning is enabled. These are the authors' recommended defaults; no threshold tuning was performed on our data.

%% file: tables/app3_tokens.tex
\begin{table}[H]
\centering
\caption{Operational 21-marker kernel, ranked by benchmark-weighted point-biserial correlation with correctness on the 115{,}200-rollout pool ($n_{\text{eff}}{=}32{,}797$ stratified). Prev.\ is rollout prevalence and W/C the wrong-to-correct frequency ratio; $^{\dagger}$marks a marker essentially absent from correct rollouts, where the ratio is unstable.}
\label{tab:tokens}
\resizebox{\columnwidth}{!}{%
\begin{tabular}{l rrrr}
\toprule
Marker & $r_{\text{pb}}^{w}$ & $p^{w}$ & Prev.\ (\%) & W/C \\
\midrule
\texttt{perhaps} & $-$0.427 & $<$1e-300 & 42.7 & 3.56$\times$ \\
\texttt{Wait,} & $-$0.286 & $<$1e-300 & 97.3 & 1.83$\times$ \\
\texttt{Wait} & $-$0.276 & $<$1e-300 & 97.7 & 1.79$\times$ \\
\texttt{actually} & $-$0.210 & $<$1e-300 & 60.4 & 1.75$\times$ \\
\texttt{wrong} & $-$0.142 & $<$1e-300 & 23.4 & 2.32$\times$ \\
\texttt{However} & $-$0.125 & $<$1e-300 & 18.0 & 1.89$\times$ \\
\texttt{Or } & $-$0.122 & $<$1e-300 & 12.9 & 2.36$\times$ \\
\texttt{no,} & $-$0.108 & $<$1e-300 & 58.5 & 1.80$\times$ \\
\texttt{Alternatively} & $-$0.094 & $<$1e-300 & 64.3 & 1.28$\times$ \\
\texttt{But wait} & $-$0.090 & $<$1e-300 & 43.3 & 1.47$\times$ \\
\texttt{Actually} & $-$0.072 & $<$1e-300 & 2.7 & 2.06$\times$ \\
\texttt{Maybe} & $-$0.063 & $<$1e-300 & 66.2 & 1.13$\times$ \\
\texttt{incorrect} & $-$0.059 & $<$1e-300 & 10.2 & 1.74$\times$ \\
\texttt{instead} & $-$0.057 & $<$1e-300 & 30.7 & 1.08$\times$ \\
\texttt{I was wrong} & $-$0.045 & 2e-16 & 0.7 & 3.20$\times$ \\
\texttt{Let me think} & $-$0.032 & 7e-9 & 43.1 & 0.93$\times$ \\
\texttt{Let me reconsider} & $-$0.024 & 1e-5 & 0.0 & n/a$^{\dagger}$ \\
\texttt{I think I made} & $-$0.023 & 3e-5 & 2.2 & 2.11$\times$ \\
\texttt{Re-examining} & $-$0.017 & 0.002 & 0.0 & n/a$^{\dagger}$ \\
\texttt{This isn't} & $-$0.016 & 0.003 & 0.0 & n/a$^{\dagger}$ \\
\texttt{Let me try} & $-$0.015 & 0.008 & 43.4 & 1.08$\times$ \\
\bottomrule
\end{tabular}%
}
\end{table}

%% file: tables/app_decile_density.tex
\begin{table}[t]
\centering
\small
\setlength{\tabcolsep}{4pt}
\renewcommand{\arraystretch}{0.92}
\begin{tabular}{lcc}
\toprule
Decile & Density range & Accuracy (\%) \\
\midrule
D1 & 0.00--1.11 & 96.9 \\
D2 & 1.11--1.76 & 98.0 \\
D3 & 1.76--2.30 & 97.0 \\
D4 & 2.30--2.80 & 95.9 \\
D5 & 2.80--3.31 & 95.1 \\
D6 & 3.31--3.87 & 94.3 \\
D7 & 3.87--4.57 & 93.0 \\
D8 & 4.57--5.48 & 89.1 \\
D9 & 5.48--6.75 & 80.1 \\
D10 & $>$6.75 & 65.6 \\
\bottomrule
\end{tabular}
\caption{Accuracy by hesitation-marker density decile across 115{,}200 rollouts, where density is marker count per 1K characters. Accuracy falls monotonically with density ($r{=}-0.82$ between decile index and accuracy), for a 31.4pp gap between the lowest and highest decile.}
\label{tab:density_decile}
\end{table}

%% file: tables/app_model_density.tex
\begin{table}[t]
\centering
\resizebox{\columnwidth}{!}{%
\begin{tabular}{lcccc}
\toprule
Model & All & Correct & Wrong & Wrong/Correct \\
\midrule
DS-R1-1.5B & 4.05 & 3.12 & 7.23 & 2.32 \\
DS-R1-7B & 3.37 & 2.93 & 6.66 & 2.27 \\
OT3-7B & 5.11 & 4.98 & 6.56 & 1.32 \\
Qwen3-4B & 3.13 & 3.10 & 3.59 & 1.16 \\
Qwen3-30B & 2.44 & 2.43 & 2.55 & 1.05 \\
QwQ-32B & 4.30 & 4.17 & 6.73 & 1.62 \\
\bottomrule
\end{tabular}%
}
\caption{Per-model hesitation-marker density, in markers per 1K characters, where wrong counts every rollout not graded correct. Absolute density varies several-fold across models, but every model separates wrong from correct in the same direction, which is what lets one relative rule transfer.}
\label{tab:model_density}
\end{table}

%% file: tables/app_phi4_pathology.tex
\begin{table}[t]
\centering
\resizebox{\columnwidth}{!}{%
\begin{tabular}{lcccc}
\toprule
Benchmark & Pass@1 & SC@32 & Pass@32 & No-ans./32 \\
\midrule
AIME24 & 15.7 & 50.0 & 73.3 & 19.0 \\
AIME25 & 13.6 & 36.7 & 66.7 & 17.2 \\
AMC23 & 16.8 & 55.0 & 87.5 & 18.2 \\
MATH500 & 11.5 & 46.6 & 59.6 & 21.0 \\
\bottomrule
\end{tabular}%
}
\caption{Phi-4-reasoning rollout profile at $k{=}32$. Accuracy columns are percentages under exact-match grading (Appendix~\ref{app:generation}); no-answer reports the mean number of rollouts per problem that never produce an extractable final answer.}
\label{tab:phi4_pathology}
\end{table}

%% file: tables/app_no_answer_reduction.tex
\begin{table}[t]
\centering
\resizebox{\columnwidth}{!}{%
\begin{tabular}{lcccc}
\toprule
Subset & Start & Ckpt 1 & Ckpt 2 & Final \\
\midrule
AIME/AMC avg. & 2.67 & 1.74 & 1.18 & 0.79 \\
Qwen3-4B/AIME25 & 9.4 & -- & -- & 2.4 \\
OT3-7B/AIME25 & 5.4 & -- & -- & 1.7 \\
\bottomrule
\end{tabular}%
}
\caption{No-answer rollouts remaining under FoT over the checkpoint sequence on the AIME24/AIME25/AMC23 stress subset. Entries report the mean count per 32-rollout pool; the aggregate removes 71\% of no-answer rollouts by the final checkpoint.}
\label{tab:no_answer_reduction}
\end{table}

%% file: tables/app1_gap.tex
\begin{table}[H]
\centering
\renewcommand{\arraystretch}{0.92}
\resizebox{\columnwidth}{!}{
    \begin{tabular}{ll cccc}
    \toprule
    & & \multicolumn{4}{c}{\textbf{Accuracy (\%)}} \\
    \cmidrule(lr){3-6}
    Model & Dataset & Pass@1 & Pass@32 & Gap$_{1 \to 32}$ & SC@32 \\
    \midrule
    \multirow{4}{*}{DS-R1-1.5B}
      & AIME24  & 29.2 & 80.0 & 50.8 & 56.7 \\
      & AIME25  & 24.3 & 53.3 & 29.1 & 33.3 \\
      & AMC23   & 71.2 & 100.0 & 28.8 & 90.0 \\
      & MATH500 & 83.8 & 98.0 & 14.2 & 92.4 \\[2pt]
    \multirow{4}{*}{DS-R1-7B}
      & AIME24  & 54.9 & 83.3 & 28.4 & 80.0 \\
      & AIME25  & 39.1 & 66.7 & 27.6 & 53.3 \\
      & AMC23   & 90.5 & 100.0 & 9.5 & 95.0 \\
      & MATH500 & 93.2 & 99.0 & 5.8 & 95.4 \\[2pt]
    \multirow{4}{*}{OT3-7B}
      & AIME24  & 63.6 & 90.0 & 26.4 & 76.7 \\
      & AIME25  & 60.7 & 80.0 & 19.3 & 73.3 \\
      & AMC23   & 93.1 & 100.0 & 6.9 & 97.5 \\
      & MATH500 & 95.2 & 99.0 & 3.8 & 97.2 \\[2pt]
    \multirow{4}{*}{Qwen3-4B}
      & AIME24  & 70.6 & 90.0 & 19.4 & 86.7 \\
      & AIME25  & 68.8 & 86.7 & 17.9 & 86.7 \\
      & AMC23   & 99.0 & 100.0 & 1.0 & 100.0 \\
      & MATH500 & 97.4 & 99.4 & 2.0 & 98.4 \\[2pt]
    \multirow{4}{*}{Qwen3-30B}
      & AIME24  & 86.8 & 93.3 & 6.6 & 93.3 \\
      & AIME25  & 80.9 & 90.0 & 9.1 & 86.7 \\
      & AMC23   & 99.7 & 100.0 & 0.3 & 100.0 \\
      & MATH500 & 97.4 & 99.6 & 2.2 & 98.0 \\[2pt]
    \multirow{4}{*}{QwQ-32B}
      & AIME24  & 80.2 & 93.3 & 13.1 & 90.0 \\
      & AIME25  & 69.7 & 90.0 & 20.3 & 83.3 \\
      & AMC23   & 98.2 & 100.0 & 1.8 & 100.0 \\
      & MATH500 & 96.8 & 99.0 & 2.2 & 97.4 \\[2pt]
    \midrule
    \multicolumn{2}{l}{\textbf{Average}} & 76.8 & 91.3 & 14.4 & 85.9 \\
    \bottomrule
    \end{tabular}
}
\caption{Pass@1 vs.\ Pass@32 accuracy gap across 6 models and 4 benchmarks ($k{=}32$), where Pass@1 is the average single-rollout accuracy and Pass@32 the oracle accuracy of the pool. The 14.4pp average gap is the headroom majority voting exploits; averages are unweighted means over the 24 model--benchmark cells.}
\label{tab:passk_gap}

\end{table}

%% file: tables/table_stratification.tex
\begin{table}[t]
\centering
\scriptsize
\setlength{\tabcolsep}{3.5pt}
\renewcommand{\arraystretch}{0.9}
\resizebox{\columnwidth}{!}{%
\begin{tabular}{l rccc rccc}
\toprule
& \multicolumn{4}{c}{\textbf{All (4 benchmarks)}} & \multicolumn{4}{c}{\textbf{Hard (AIME24/25)}} \\
\cmidrule(lr){2-5} \cmidrule(lr){6-9}
pass@1 bin & $n$ & SC & FoT & $\Delta$ & $n$ & SC & FoT & $\Delta$ \\
\midrule
$p_1{=}0$              & 91   & 0.0   & 0.0  & $+$0.0 & 61  & 0.0   & 0.0  & $+$0.0 \\
$0{<}p_1{\leq}0.25$    & 121  & 23.1  & 28.9 & $+$5.8 & 42  & 35.7  & 40.5 & $+$4.8 \\
$0.25{<}p_1{\leq}0.5$  & 91   & 79.1  & 80.2 & $+$1.1 & 33  & 93.9  & 97.0 & $+$3.0 \\
$p_1{>}0.5$            & 3297 & 100.0 & 99.9 & $-$0.1 & 224 & 100.0 & 99.1 & $-$0.9 \\
\bottomrule
\end{tabular}%
}
\caption{Accuracy (\%) stratified by per-problem difficulty, graded with \texttt{math\_verify}. Bins are per-problem pass@1, the fraction of the 32 rollouts that are correct; $n$ counts model-problem pairs.}
\label{tab:stratification}
\end{table}

%% file: tables/table_baselines.tex
\begin{table}[t]
\centering

\resizebox{\columnwidth}{!}{%
\setlength{\tabcolsep}{5pt}
\renewcommand{\arraystretch}{0.95}
\begin{tabular}{l cc cc}
\toprule
& \multicolumn{2}{c}{\textbf{All}} & \multicolumn{2}{c}{\textbf{Hard}} \\
\cmidrule(lr){2-3} \cmidrule(lr){4-5}
Method & Acc & FLOP$\downarrow$ & Acc & FLOP$\downarrow$ \\
\midrule
SC@32                        & 94.4 & ---            & 75.0 & ---           \\
DSC                          & 94.4 & 69.4\%         & 75.0 & 54.4\%        \\
Certaindex (offline proxy)   & 94.0 & 11.5--13.2\%   & 75.0 & 9.9--10.9\%   \\
\textbf{FoT@32}              & 94.5 & 28.8\%         & 75.0 & 44.0\%        \\
FoT+DSC (composed)           & 94.0 & 80.9\%         & 73.6 & 75.7\%        \\
\bottomrule
\end{tabular}%
}

\caption{Sample-axis baselines and composition on a unified basis: \texttt{math\_verify} grading, with FLOP$\downarrow$ the \emph{full-model} saving against SC@32 and hard = AIME24/25. DSC \citep{wang-etal-2025-make} and Certaindex \citep{fu2025efficientlyscalingllmreasoning} act on the sample axis, FoT on the token axis.}
\label{tab:baselines}
\end{table}

%% file: tables/table6_ablation.tex
\begin{table}[H]
\centering

\resizebox{\columnwidth}{!}{%
\begin{tabular}{l ccc}
\toprule
& \multicolumn{3}{c}{Keep ratio $\rho$} \\
\cmidrule(lr){2-4}
Schedule & 0.50 & 0.67 & 0.75 \\
\midrule
Early \{4K, 8K, 12K\}  & 3380 ($-$17) & 3398 ($+$1) & 3400 ($+$3) \\
Mid \{6K, 10K, 14K\}  & 3389 ($-$8) & \cellcolor{gray!20}\textbf{3401 ($+$4)} & 3403 ($+$6) \\
Math-only \{6K, 11K, 16K\}  & 3399 ($+$2) & 3404 ($+$7) & 3399 ($+$2) \\
Late \{8K, 12K, 16K\}  & 3395 ($-$2) & 3401 ($+$4) & 3402 ($+$5) \\
\bottomrule
\end{tabular}%
}

\caption{Ablation over checkpoint schedules and keep ratio $\rho$ on the 4-benchmark pool (3{,}600 pairs per cell), showing FoT correct count and $\Delta$ against the SC@32 baseline of 3397/3600. The shaded universal default (Mid \{6K, 10K, 14K\}, $\rho{=}0.67$) is our joint choice; within this grid Math-only \{6K, 11K, 16K\} scores highest in domain but loses out of distribution (Appendix~\ref{app:sensitivity}).}
\label{tab:ablation}
\end{table}

%% file: tables/table8_permodel.tex
\begin{table}[H]

\centering
\setlength{\tabcolsep}{4pt}
\renewcommand{\arraystretch}{0.92}
\small
\resizebox{\columnwidth}{!}{%
\begin{tabular}{l l cc cc}
\toprule
& & \multicolumn{2}{c}{\textbf{Per-Model}} & \multicolumn{2}{c}{\textbf{Universal}} \\
\cmidrule(lr){3-4} \cmidrule(lr){5-6}
Model & Best Config & $\Delta$ & FLOP$\downarrow$ & $\Delta$ & FLOP$\downarrow$ \\
\midrule
DS-R1-1.5B   & Math-only, $\rho{=}0.50$ & $+$7 & 63.2 & $+$2 & 52.7 \\
DS-R1-7B     & Math-only, $\rho{=}0.50$ & $+$4 & 55.5 & $+$3 & 46.5 \\
OT3-7B       & Mid, $\rho{=}0.67$ & $+$3 & 51.2 & $+$3 & 51.2 \\
Qwen3-4B     & Mid, $\rho{=}0.75$ & $-$1 & 39.6 & $-$2 & 49.1 \\
Qwen3-30B    & Math-only, $\rho{=}0.67$ & $+$2 & 43.0 & $+$0 & 45.3 \\
QwQ-32B      & Math-only, $\rho{=}0.50$ & $+$0 & 53.5 & $-$2 & 43.9 \\
\midrule
\textbf{Agg.} & & \textbf{$+$15} & \textbf{51.0} & \textbf{$+$4} & \textbf{48.1} \\
\bottomrule
\end{tabular}%
}

\caption{Per-model calibration against the universal Mid $\rho{=}0.67$ config, with each model's best config selected on its own 600-problem math pool. $\Delta$ is relative to SC@32 and Agg.\ sums across all 3{,}600 model-problem pairs.}
\label{tab:permodel}
\end{table}

%% file: tables/app2_transfer.tex
\begin{table*}[t!]
\centering

\setlength{\tabcolsep}{5pt}
\renewcommand{\arraystretch}{0.92}
\small
\begin{tabular}{l cccccc c}
\toprule
& \multicolumn{6}{c}{\textbf{Target Model ($\Delta$ vs.\ SC@32 on 600 problems)}} & \\
\cmidrule(lr){2-7}
Source Config & DS-R1-1.5B & DS-R1-7B & OT3-7B & Qwen3-4B & Qwen3-30B & QwQ-32B & Agg. \\
\midrule
Mid, $\rho{=}0.67$ (Ours) & $+$2 & $+$3 & \textbf{$+$3} & $-$2 & $+$0 & $-$2 & $+$4 \\
Math-only, $\rho{=}0.50$ & \textbf{$+$7} & \textbf{$+$4} & $-$2 & $-$6 & $-$1 & \textbf{$+$0} & $+$2 \\
Math-only, $\rho{=}0.67$ & $+$4 & $+$4 & $+$1 & $-$3 & \textbf{$+$2} & $-$1 & $+$7 \\
Mid, $\rho{=}0.75$ & $+$3 & $+$2 & $+$2 & \textbf{$-$1} & $+$1 & $-$1 & $+$6 \\
\bottomrule
\end{tabular}

\caption{Cross-model config transfer on the 4-benchmark pool, where cells show $\Delta$ against SC@32 out of 600 problems per model and Agg.\ sums across all 3{,}600 pairs. Bold marks each model's self-optimal config from Table~\ref{tab:permodel}.}
\label{tab:transfer}

\end{table*}

%% file: tables/app_kernel_sweep.tex
\begin{table}[t]
\centering
\resizebox{\columnwidth}{!}{%
\begin{tabular}{l r r r}
\toprule
Kernel & $\Delta$ & Tok$\downarrow$ (\%) & FLOP$\downarrow$ (\%) \\
\midrule
\texttt{top-1}  (perhaps)               & $-3$ & 29.12 & 50.59 \\
\texttt{top-2}                          & $+6$ & 29.20 & 50.66 \\
\texttt{top-3}  (3 markers)              & $+5$ & 29.22 & 50.70 \\
\texttt{top-5}                          & $+5$ & 29.19 & 50.65 \\
\texttt{top-7}                          & $+6$ & 29.17 & 50.59 \\
\texttt{top-10}                         & $+0$ & 29.17 & 50.61 \\
\texttt{top-15}                         & $+3$ & 29.24 & 50.76 \\
\texttt{top-21} (default)               & $+4$ & 29.23 & 50.73 \\
\bottomrule
\end{tabular}%
}
\caption{Kernel-size sweep on the 4-benchmark pool. Tokens are added by descending $|r_{\text{pb}}|$; $\Delta$ is FoT@32 minus SC@32.}
\label{tab:kernel_sweep}
\end{table}